\documentclass[lettersize,journal]{IEEEtran}
\usepackage{amsmath,amsfonts}
\usepackage{algorithmic}
\usepackage{algorithm}
\usepackage{array}
\usepackage{textcomp}
\usepackage{stfloats}
\usepackage{url}
\usepackage{verbatim}
\usepackage{graphicx}
\usepackage{cite}
\usepackage{subfigure}
\usepackage{graphicx}
\usepackage{amsmath} 
\usepackage{multirow}
\usepackage{caption}
\usepackage[table]{xcolor}
\usepackage{arydshln}

\begin{document}

\title{Contrastive Representation Distillation via Multi-Scale Feature Decoupling}

\author{Cuipeng Wang,~\IEEEmembership{Student Member,~IEEE,}~Haipeng~Wang,~\IEEEmembership{Senior Member,~IEEE}

\thanks{This work was supported by the National Natural Science Foundation of China (Grant No. 62271153). (Corresponding author: Haipeng Wang).}
\thanks{Cuipeng Wang, Haipeng Wang are with the Key Laboratory of Information Science of Electromagnetic Waves, Fudan University, Shanghai 200433, China (e-mail: cpwang23@m.fudan.edu.cn, hpwang@fudan.edu.cn).}
}

\markboth{Journal of \LaTeX\ Class Files,~Vol.~14, No.~8, August~2021}%
{Shell \MakeLowercase{\textit{et al.}}: A Sample Article Using IEEEtran.cls for IEEE Journals}


\maketitle

\begin{abstract}
Knowledge distillation enhances the performance of compact student networks by transferring knowledge from more powerful teacher networks without introducing additional parameters.
In the feature space, local regions within an individual global feature encode distinct yet interdependent semantic information.
Previous feature-based distillation methods mainly emphasize global feature alignment while neglecting the decoupling of local regions within an individual global feature, which often results in semantic confusion and suboptimal performance.
Moreover, conventional contrastive representation distillation suffers from low efficiency due to its reliance on a large memory buffer to store feature samples.
To address these limitations, this work proposes MSDCRD, a model-agnostic distillation framework that systematically decouple global features into multi-scale local features and leverages the resulting semantically rich feature samples with tailored sample-wise and feature-wise contrastive losses.
This design enables efficient distillation using only a single batch, eliminating the dependence on external memory.
Extensive experiments demonstrate that MSDCRD achieves superior performance not only in homogeneous teacher–student settings but also in heterogeneous architectures where feature discrepancies are more pronounced, highlighting its strong generalization capability.
\end{abstract}


\section{Introduction}
\label{Introduction}
\IEEEPARstart{T}{he} past few decades have witnessed remarkable achievements of neural networks in the field of computer vision. 
With the continuous increase in network depth and width~\cite{he2016deep}, model performance has also been significantly improved.
Deeper networks with more parameters have brought improved performance; however, they also come with trade-offs.
Deeper networks incur higher computational and memory overhead, posing significant challenges for deployment on resource-limited devices.
To address this issue, various model compression techniques have been proposed, including model pruning~\cite{frankle2018lottery,li2016pruning,liu2018rethinking,luo2017thinet}, model quantization~\cite{jacob2018quantization,courbariaux2015binaryconnect}, lightweight network design~\cite{howard2017mobilenets,sandler2018mobilenetv2,zhang2018shufflenet}, and knowledge distillation~\cite{hinton2015distilling,zagoruyko2016paying,romero2014fitnets}.

Knowledge distillation(KD) is a specialized form of transfer learning, where the ``knowledge'' from a large pre-trained network (also known as teacher) is transferred to a small student network (a.k.a. student) to enhance the performance of the latter.
Moreover, KD can be seamlessly combined with other model compression techniques.
For instance, in~\cite{fu2024instance}, lightweight networks are integrated with Density Map Knowledge Distillation to address the issue of accuracy degradation in compact models.
Given its capability to further improve network performance, KD has emerged as a compelling and enduring focus in the field of efficient deep neural networks.
\begin{figure}[t]
\centering
\includegraphics[width=3.5in]{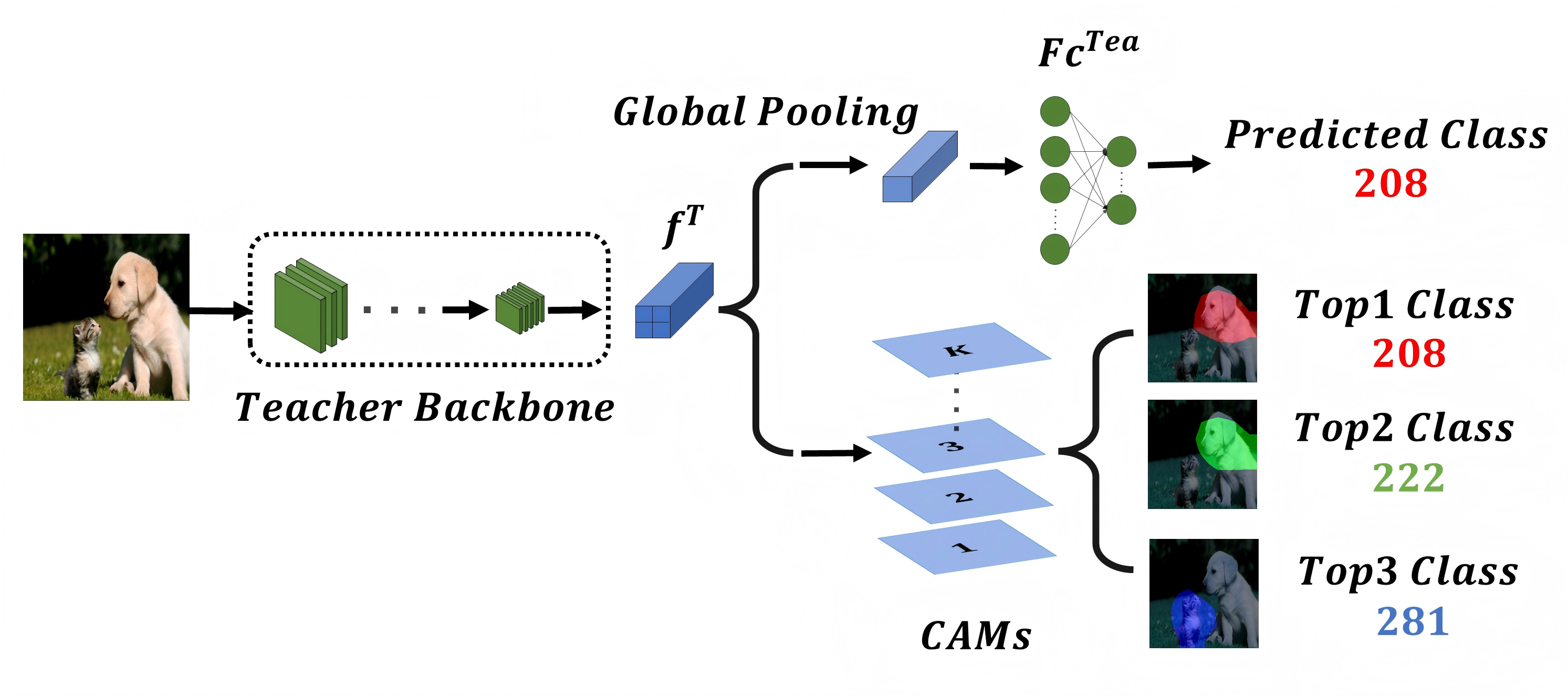}
\caption{\textbf{Different Local Regions in The CAMs Focus on Distinct Category Information.} Different colors denote different classes.}
\label{CAMs}
\vskip -0.25in
\end{figure}
\begin{figure*}[t]
    \centering
    \includegraphics[width=\textwidth]{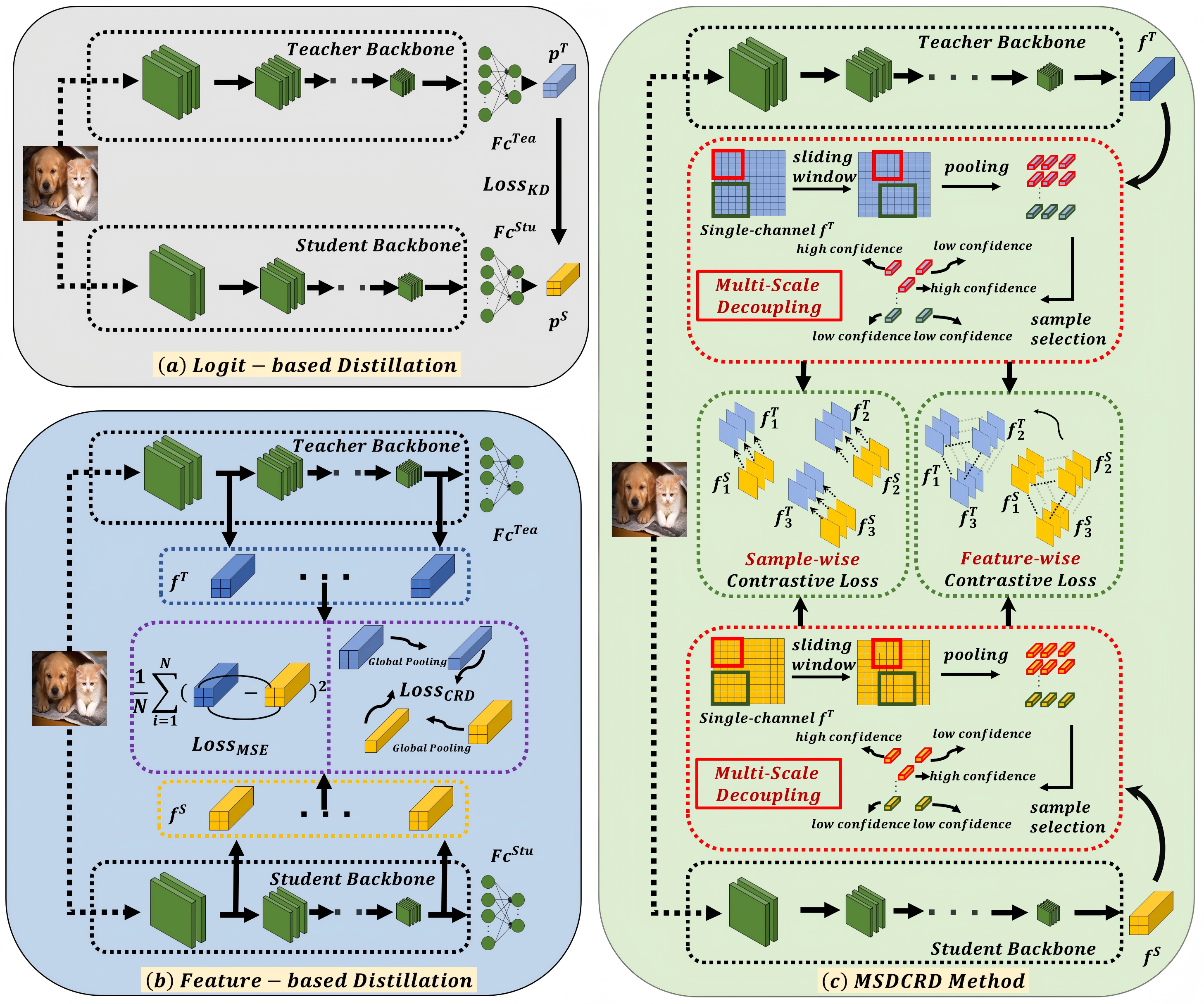}
\caption{\textbf{Illustration of Previous Distillation Methods and The Proposed MSDCRD.} 
(a) Logit-based distillation, which minimizes the KL divergence between the student and teacher logits.
(b) Previous feature-based distillation, which align the global feature representations of the student and teacher networks.
(c)The proposed MSDCRD method, which performs multi-scale decoupling on an individual global feature and integrates the resulting features with two efficient sample-wise and feature-wise contrastive losses.
}
\vskip -0.2in
\label{Figure_Process}
\end{figure*}

Hinton et al.~\cite{hinton2015distilling} first propose distilling a teacher's knowledge into a student by minimizing the Kullback-Leibler (KL) divergence between their logits.
Building on this foundation, a series of subsequent works have extended logit-based KD.
For example, DKD~\cite{zhao2022decoupled} decouples the knowledge in logits into target-class and non-target-class information for transfer.
CKD~\cite{zhang2024collaborative}, building upon DKD, analyzes that the teacher’s overconfidence suppresses the transfer of dark knowledge, thereby hindering the student’s learning potential.
To address this, a feature fusion module is introduced to allow the student to participate in the process of knowledge refinement, collaborating with the teacher to jointly generate knowledge more suitable for the student.
ReKD~\cite{xu2023improving} decouples logit information into head categories knowledge and tail categories knowledge, thereby exploiting both target–similar class relations and improving the utilization of tail categories knowledge.
However, logit-based knowledge distillation methods fail to exploit the rich knowledge embedded in the intermediate layers of the teacher.
To overcome this limitation, FitNet~\cite{romero2014fitnets} firstly introduces distillation using intermediate feature, opening a new line of research beyond logit matching.
Following this direction, many feature-based distillation methods~\cite{tian2019contrastive,chen2022knowledge,chen2021distilling,heo2019comprehensive,park2019relational,ahn2019variational,passalis2020probabilistic,zhou2025all,wang2024improving,deng2022distpro,liu2023norm,huang2022knowledge} have been developed.
In subsequent research, many notable works have explored efficient knowledge distillation from perspectives distinct from feature-based approaches.
For instance, AT~\cite{zagoruyko2016paying} introduces attention-based distillation by transferring knowledge through attention maps.
SKD~\cite{yang2023skill} extends conventional KD by incorporating two meta-learning networks, Teacher Behavior Teaching and Teacher Experience Teaching, enabling the student to learn not only the teacher’s knowledge but also its behavioral patterns and experiential strategies.
EKD\cite{zhang2021student} breaks the limitation of a fixed teacher by jointly training the teacher and student to progressively narrow the capacity gap, while introducing Within-Stream and Cross-Stream distillation paths for enhanced knowledge transfer.
DCGD~\cite{xu2023learning} proposes a reflective distillation paradigm by introducing the Mutual Error Distance (MED) to quantify the sufficiency of the decision boundary, revealing teacher confusion patterns, and employing divide-and-conquer group distillation to allow the student to learn both the teacher’s successes and failures.

When reviewing existing feature-based distillation methods, it is evident that they typically align teacher and student networks in the feature space at a global level.
However, as illustrated in Figure~\ref{CAMs}, Class Activation Map (CAM) visualizations indicate that different local regions within an individual global feature encode diverse high-dimensional semantics.
Some local features may encode category information that diverges from the global feature, as illustrated by the blue-masked region in the CAMs.
Furthermore, within the features corresponding to the same target, the contributions of different local regions to category discrimination are not uniform, for example, the red-masked region in the CAMs emphasizes the head of a dog, whereas the body region receives comparatively less attention.
Moreover, these local regions often exhibit interdependencies, and such internal coupling introduces additional challenges for achieving effective fine-grained feature distillation.
Consequently, methods relying solely on global feature alignment fail to exploit fine-grained local information and inter-regional couplings, preventing the student from fully capturing the teacher’s knowledge and leading to suboptimal performance.

To address the aforementioned challenges, this work proposes a novel method that performs multi-scale decoupling within individual feature maps.
Concretely, this work introduces a multi-scale sliding window pooling mechanism combined with sapmle selection, which enables fine-grained semantic decoupling  of different local regions within an individual global feature.
This design enables the generation of multiple decoupled local feature representations at varying scales, effectively capturing fine-grained local information that is typically neglected in previous approaches.
In addition, conventional CRD~\cite{tian2019contrastive} depends on complex sample processing and large memory buffers to store teacher and student features for constructing sufficient negative samples.
Such strategies not only introduce substantial computational burden but also significantly increase storage requirements, thereby limiting their practicality and scalability.
This work integrates multi-scale feature decoupling with contrastive representation distillation, enabling highly efficient contrastive representation distillation using only a single batch of features.
By leveraging the rich fine-grained feature samples obtained through multi-scale decoupling, contrastive pairs are constructed not only across different samples within a single batch but also across distinct local regions within an individual global feature.
This substantially enhances the diversity of positive and negative pairs without relying on an additional large memory buffer.
This work also analyzes the limitations of the conventional contrastive loss and accordingly proposes two complementary objectives, which are respectively termed sample-wise contrastive loss and feature-wise contrastive loss, thereby facilitating more efficient feature transfer.
The overall framework is termed Contrastive Representation Distillation via Multi-Scale Decoupling (\textbf{MSDCRD}). 
It is worth noting that the proposed method is parameter-free and can be seamlessly applied in a plug-and-play manner.

Most existing feature-based distillation methods primarily focus on knowledge transfer between homogeneous models~\cite{romero2014fitnets,tian2019contrastive,chen2022knowledge,heo2019comprehensive,ahn2019variational,passalis2020probabilistic, wang2024improving, liu2023norm, deng2022distpro}, with only a few addressing the more challenging heterogeneous models~\cite{hao2023one,zhou2025all}.
This is primarily due to the inherent differences in feature representations across different networks, which become even more pronounced when transferring knowledge between heterogeneous models.
In Section~\ref{FRG}, this work introduces Centered Kernel Alignment(\textbf{CKA}) as a similarity metric to explicitly quantify these feature discrepancies across different models.
It is worth noting that MSDCRD not only achieves outstanding performance in homogeneous models but also demonstrates strong effectiveness across diverse heterogeneous models.

In summary, the main contributions of this work are as follows:
\begin{itemize}
\item This work identifies that existing feature distillation methods often overlook the interdependencies among local regions within individual feature maps, leading to suboptimal knowledge transfer. A multi-scale feature decoupling mechanism is introduced to decouple the interdependencies among local regions within individual feature maps, enabling the student to more effectively learn fine-grained knowledge from the teacher.
\item This work proposes a model-agnostic method, \textbf{MSDCRD}, which integrates multi-scale feature decoupling with contrastive learning to eliminate the reliance of prior contrastive representation distillation on complex data processing and large memory buffers. In addition, it incorporates two tailored contrastive loss functions that further enhance the efficiency of knowledge transfer.
\item Extensive experiments are conducted on several visual benchmarks across both homogeneous and heterogeneous models. MSDCRD consistently enables significant performance improvements of student models across a variety of teacher-student architecture pairs, demonstrating its superior effectiveness.
\end{itemize}
\section{Related Work}
\label{Related_Work}
Knowledge distillation transfers the knowledge of a complex teacher network to a lightweight student network, enhancing the performance of the student network.
Knowledge distillation was first proposed by Hinton et al.~\cite{hinton2015distilling}, subsequent studies~\cite{luo2024scale, zhao2022decoupled, sun2024logit,jin2023multi,li2023curriculum, wang2025abkd, xu2023improving, zhang2024collaborative} have focused on developing diverse methodological advancements in the logit space for more effective knowledge transfer.
Due to the limitations of logit-space distillation and its inability to fully exploit the intermediate feature knowledge of the teacher network, FitNet~\cite{romero2014fitnets} first proposed distillation in the feature space.
Since then, an increasing number of studies have focused on intermediate feature-based distillation~\cite{tian2019contrastive,chen2022knowledge,chen2021distilling,heo2019comprehensive,park2019relational,ahn2019variational,passalis2020probabilistic,zhou2025all,wang2024improving,hao2023one,huang2022knowledge}.
Meanwhile, AT~\cite{zagoruyko2016paying} was the first to propose distillation using attention maps derived from intermediate features, and subsequently, CAT-KD~\cite{guo2023class} leveraged class activation maps for knowledge transfer.

\textbf{Feature-Based Knowledge Distillation in Homogeneous Models.}
Due to the strong task generality of feature-based distillation, an increasing number of researchers have proposed enhancements building upon the foundation established by FitNet~\cite{romero2014fitnets}.
PKT~\cite{passalis2020probabilistic} aligned the probability distributions of teacher and student network features by minimizing their statistical divergence.
OFD~\cite{heo2019comprehensive} distilled the features immediately preceding the final activation function at each stage of the network, and further proposed a novel activation function and a partial L2 loss to boost the performance of the student network.
CRD~\cite{tian2019contrastive} achieved efficient feature transfer by integrating knowledge distillation with contrastive learning.
ReviewKD~\cite{chen2021distilling} improved knowledge distillation by introducing a review mechanism, in which the lower-layer features of the teacher guide the higher-layer features of the student.
SimKD~\cite{chen2022knowledge} achieved efficient knowledge transfer by directly reusing the teacher network’s classification layer as the student’s classifier.
Norm~\cite{liu2023norm} achieved an N-to-one alignment by mapping the student network features to N times the dimensionality of the teacher network features.

\textbf{Feature-Based Knowledge Distillation in Heterogeneous Models.}
In the past, CNN dominated as the mainstream architecture for visual tasks. 
However, models based on Transformer architectures~\cite{dosovitskiy2020image,liu2021swin,touvron2021training} and MLP-based networks~\cite{touvron2022resmlp,tolstikhin2021mlp} have recently emerged, demonstrating superior performance across various vision benchmarks.
Due to the distinct inductive biases inherent to heterogeneous models, the feature representations produced by heterogeneous models also exhibit fundamental differences.
As a result, knowledge transfer across heterogeneous models presents a more pronounced challenge compared to the case of homogeneous models.
Most of the aforementioned feature-based distillation methods are primarily designed for homogeneous architectures, which may hinder their effectiveness when the teacher and student are heterogeneous models.
To address this, OFAKD~\cite{hao2023one} transfers knowledge by mapping multi-stage student features into the logit space and computing the loss with respect to the teacher’s logits.
TCS~\cite{zhou2025all}, on the other hand, aligns student and teacher features by projecting them onto the principal component directions obtained from the teacher’s features.
Although there have been some studies that focus on feature-based distillation between heterogeneous models, the development of efficient, model-agnostic feature-based distillation techniques remains an open problem.

Existing feature-based distillation methods often overlook the significant semantic discrepancies among different local regions within an individual global feature representation, as well as the coupling relationships between these local regions. 
To address this limitation, this work introduces, for the first time, a multi-scale decoupling mechanism in the feature space, which decouples interdependencies across local regions within an individual global feature to obtain fine-grained representations for more effective knowledge transfer.
Meanwhile, given that CRD~\cite{tian2019contrastive} depends on complex data processing pipelines and large memory buffers, leading to reduced efficiency and performance, this work integrates the rich fine-grained feature samples obtained through multi-scale decoupling with contrastive learning.
As a result, it eliminates the need for such costly operations and achieves efficient contrastive representation distillation using only a single batch of samples. 
Furthermore, two novel objectives, a sample-wise contrastive loss and a feature-wise contrastive loss, are introduced to further enhance feature transfer.
MSDCRD is a parameter-free, model-agnostic contrastive representation distillation method that achieves efficient knowledge transfer using only a single batch of samples.
Extensive experiments on various visual benchmarks, covering both homogeneous and heterogeneous teacher–student settings, demonstrate the superior knowledge transfer capability of MSDCRD.
\begin{figure}[t]  
    \centering

    \subfigure[$x/y$-axis: CNN/CNN]{
        \includegraphics[width=0.45\linewidth]{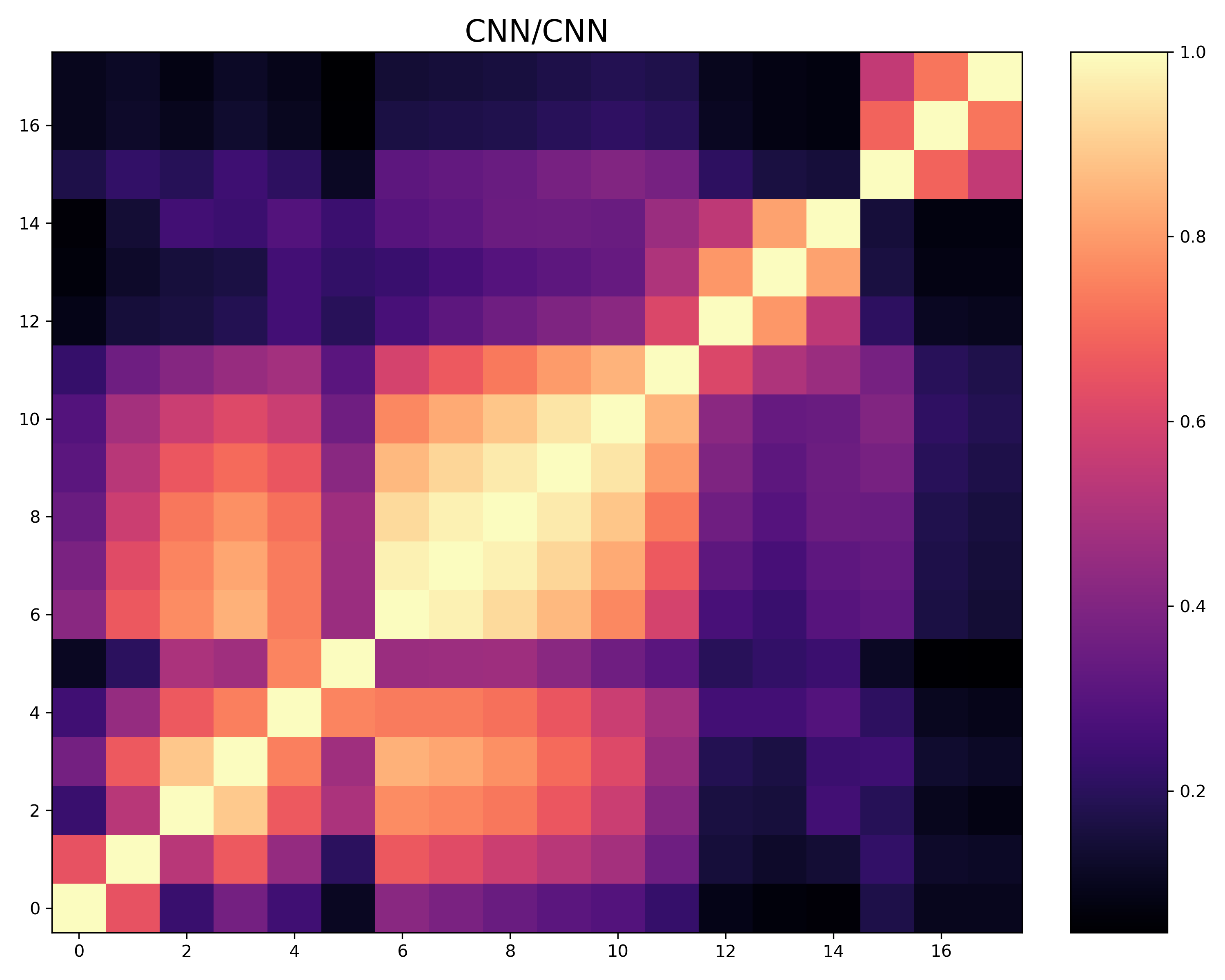}
    }
    \hfill
    \subfigure[$x/y$-axis: ViT/ViT]{
        \includegraphics[width=0.45\linewidth]{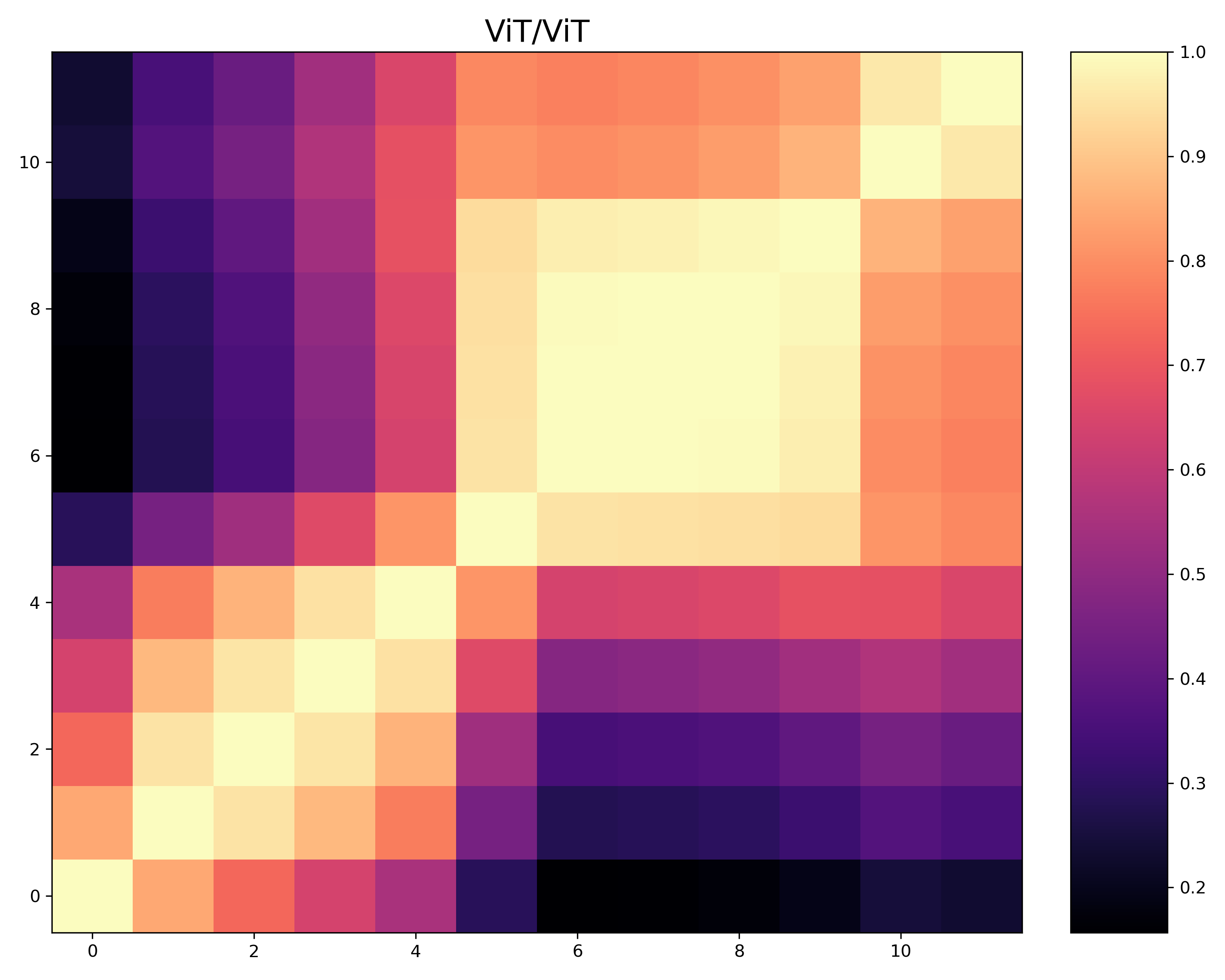}
    }

    \subfigure[$x/y$-axis: MLP/MLP]{
        \includegraphics[width=0.45\linewidth]{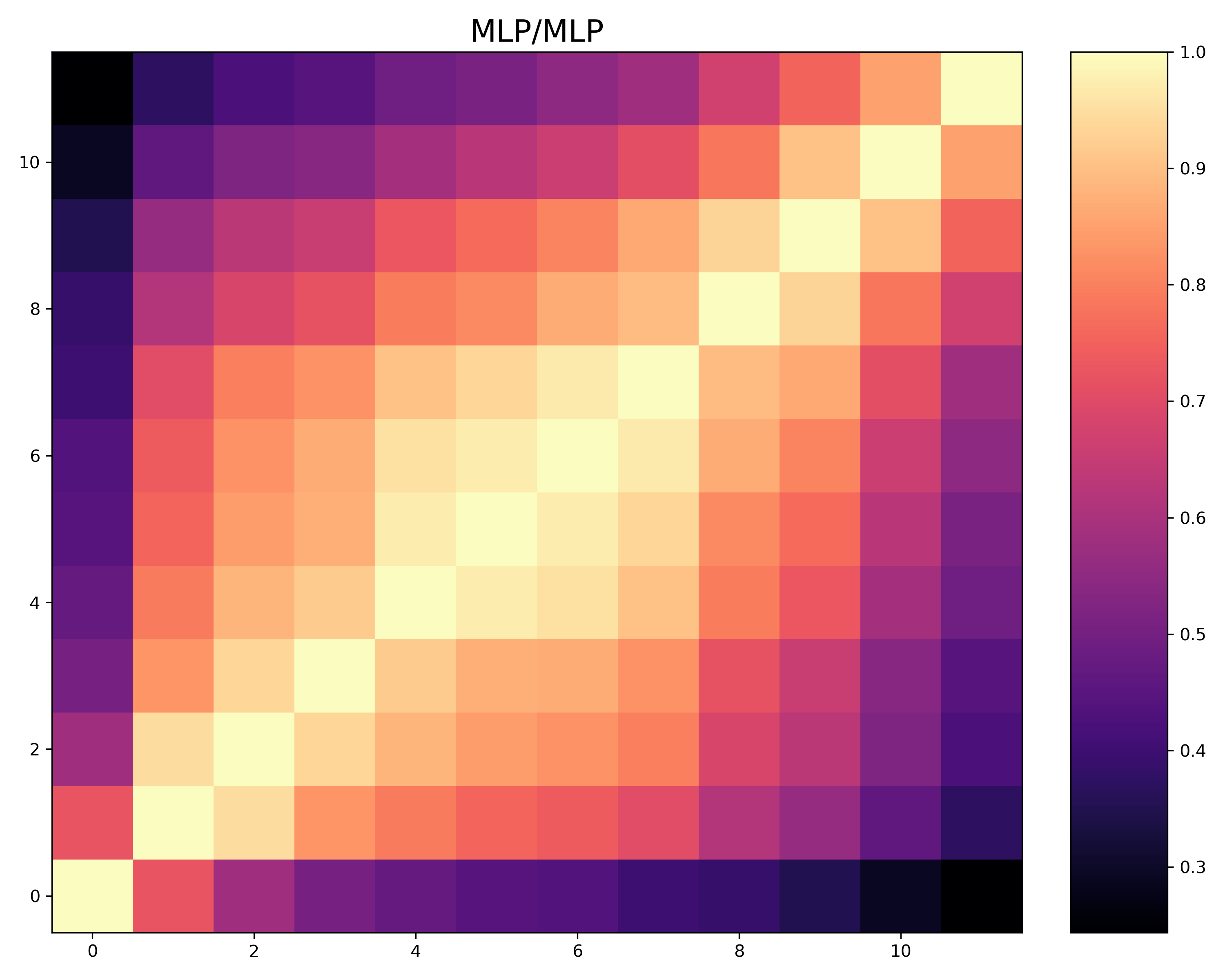}
    }
    \hfill
    \subfigure[$x/y$-axis: CNN/ViT]{
        \includegraphics[width=0.45\linewidth]{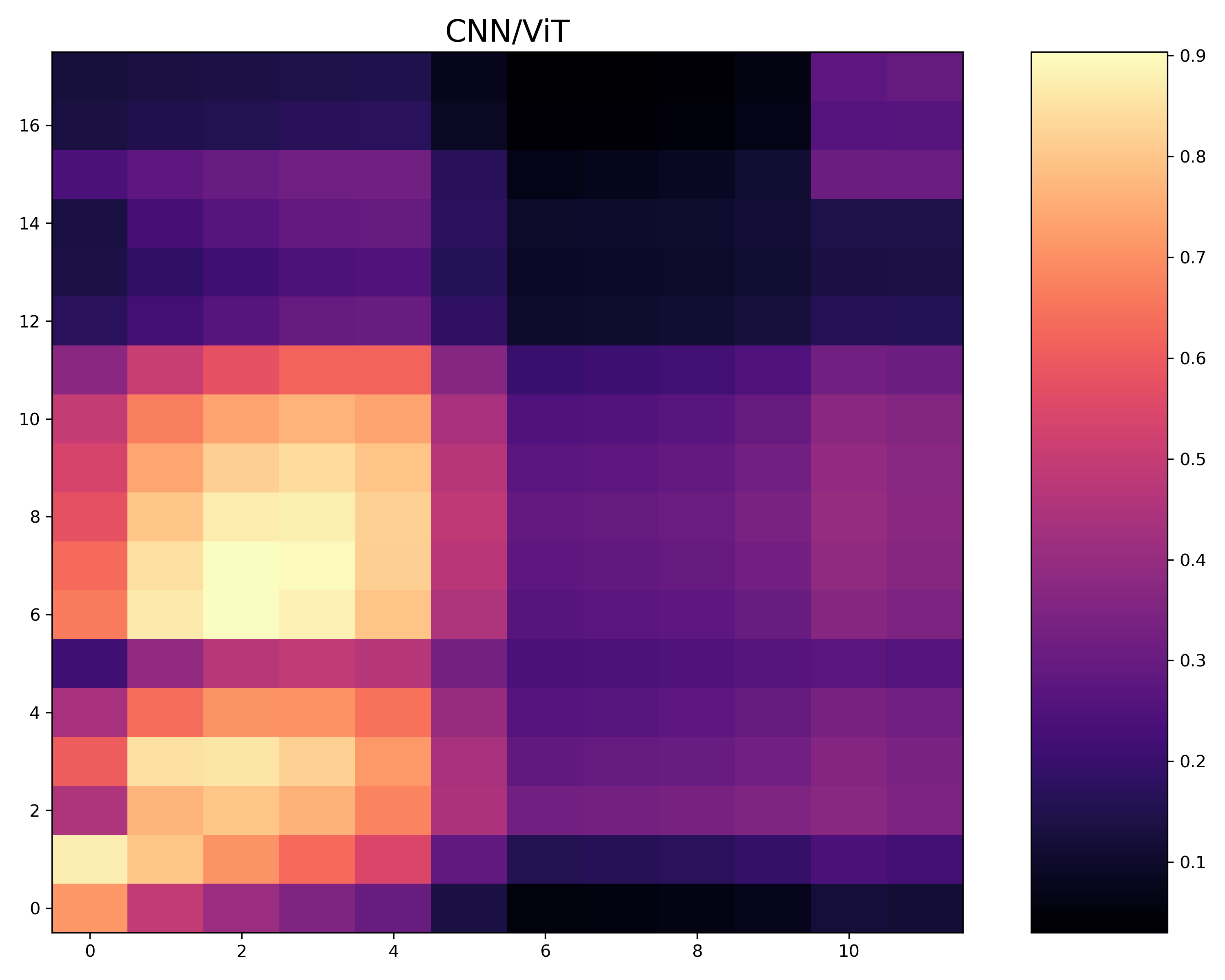}
    }

    \subfigure[$x/y$-axis: CNN/MLP]{
        \includegraphics[width=0.45\linewidth]{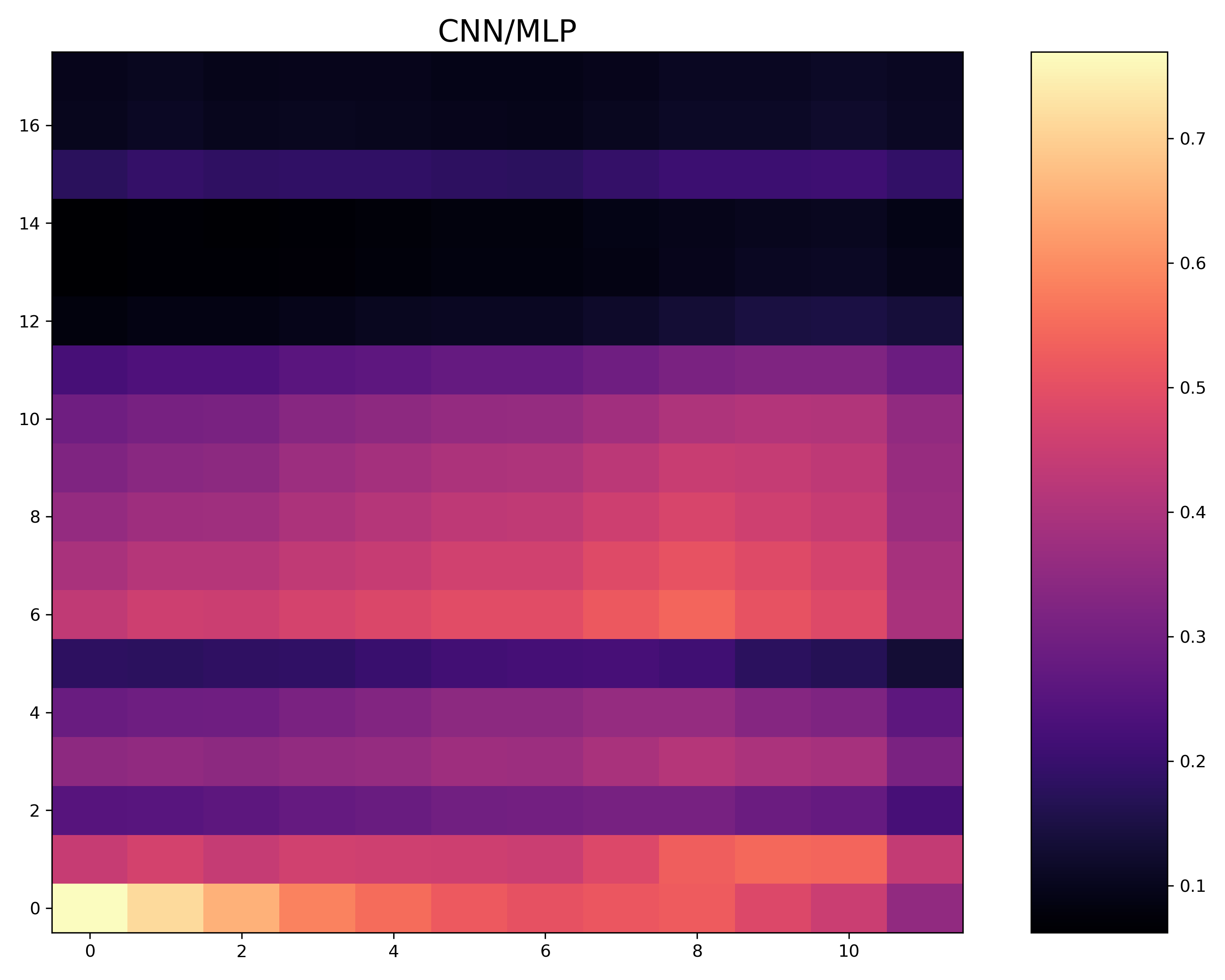}
    }
    \hfill
    \subfigure[$x/y$-axis: MLP/ViT]{
        \includegraphics[width=0.45\linewidth]{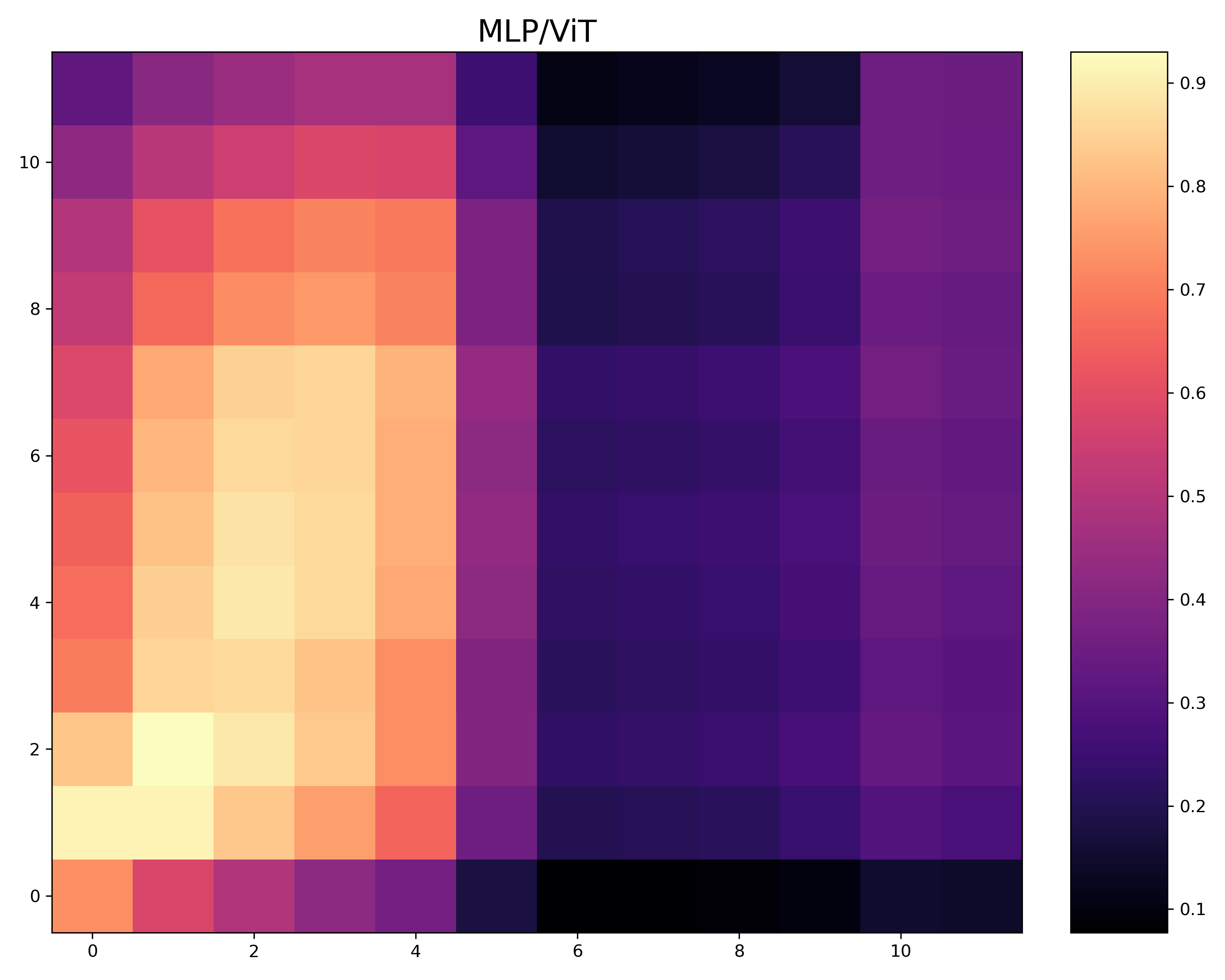}
    }

    \caption{\textbf{Similarity Heatmap of Intermediate Features in Homogeneous vs. Heterogeneous Models Measured by CKA.} This work compares features from ConvNeXt(CNN), ViT(Transformer) and Mixer(MLP).}
    \label{fig3}
\vskip -0.2in
\end{figure}
\section{Method}
\label{Method}
\subsection{Feature Representation Gaps in Homogeneous vs. Heterogeneous Models}
\label{FRG}
Before describing the proposed method, this work conducts a quantitative analysis of feature discrepancies across homogeneous architectures and across heterogeneous architectures.
Understanding these discrepancies is essential for designing effective feature-based distillation methods, as it reveals how architectural variations influence the structure of the feature space.
To explicitly quantify the differences in feature representations across distinct model architectures, this work adopts the Centered Kernel Alignment (CKA)~\cite{cortes2012algorithms,kornblith2019similarity} as the similarity metric.
Compared to other similarity measures such as cosine similarity, SVCCA, and PWCCA, CKA demonstrates superior stability and has been shown to perform more reliably in cross-architecture feature comparisons.

\textbf{Centered kernel alignment analysis.}
CKA evaluates feature similarity over a single batch.
Given the output features of two different architectures on the same batch of input samples, denoted as $X \in  R ^ {~n \times d_{1}}$, $Y \in  R ^ {~n \times d_{2}}$, where 
$n$ is the number of samples in the batch and, without loss of generality, let $d_{1} \neq d_{2}$, the corresponding feature Gram matrices for $X$ and $Y$ are computed as follows
\begin{equation}
\begin{split}
L = X X ^ { T }, K = Y Y ^ { T }
\label{eq1}
\end{split}
\end{equation}

According to the Hilbert-Schmidt Independence Criterion (HSIC), its empirical estimation is given by
\begin{equation}
\begin{split}
H S I C ( K , L ) = \frac { 1 } { ( n - 1 ) ^ { 2 } } t r ( K H L H )
\label{eq2}
\end{split}
\end{equation}
where $H$ is the centering matrix $H _ { n } = I _ { n } - \frac { 1 } { n } 1 1 ^ { T }$.

HSIC is not inherently invariant to isotropic scaling.
However, such invariance can be achieved by applying an appropriate normalization.
The normalized version of HSIC is referred to as Centered Kernel Alignment (CKA)
\begin{equation}
\begin{split}
C K A ( K , L ) = \frac { H S I C ( K , L ) } { \sqrt { H S I C ( K , K ) H S I C ( L , L ) } }
\label{eq3}
\end{split}
\end{equation}

As illustrated in Figure~\ref{fig3}, CKA is computed between the output features of each block in different network architectures, and the results are visualized as heatmaps.
Figure~\ref{fig3} (a)$\sim$(c) present the CKA heatmaps for homogeneous models, where strong feature similarity is observed between each block at the same position.
In contrast, Figure~\ref{fig3} (d)$\sim$(f) show the CKA similarity between heterogeneous models, revealing a markedly different pattern: while some similarity may still exist in the early layers, deeper blocks, which typically capture higher-level semantic representations, exhibit little to no correlation.
This observation highlights the intrinsic feature gap across heterogeneous models.

The inherent discrepancy in intermediate features across heterogeneous architectures primarily arises from the differences in the architectures' inductive biases.
In CNNs, inductive biases such as locality, two-dimensional spatial structure, and translation equivariance are embedded throughout the whole network.
In contrast, ViTs incorporate global self-attention layers that lack these biases, with only MLP sublayers exhibiting locality and translational equivariance.
Therefore, methods designed for homogeneous models may fail to achieve efficient knowledge transfer in heterogeneous settings.
How to design a model-agnostic distillation approach capable of facilitating effective knowledge transfer across both homogeneous and heterogeneous scenarios remains a fundamental challenge in feature-based distillation.
\subsection{Overview and Preparation}
\label{Overview and Preparation}
\label{op}
As illustrated in Figure~\ref{Figure_Process}, in contrast to previous feature-based distillation methods that directly align the global features of teacher and student networks, the proposed MSDCRD framework introduces a multi-scale feature decoupling mechanism that explicitly decouples semantic dependencies among local regions within an individual global feature representation.
The resulting decoupled feature sample pairs are then fed into the tailored sample-wise and feature-wise contrastive loss functions to achieve efficient knowledge transfer.

\textbf{Notation.} Given a batch of input images of size $B$, denoted as $x_{ i }$, $x_{ j }$  (where $i , j = 1 , 2 , . . . , B$), with $ i = j$ indicating the same input image, let $f^{ S }$ and
$f^{ T }$ are the feature extractors of the student and teacher networks, respectively.
Let the penultimate layer feature of the teacher network be represented as $ T ^ { j } = f ^ { T } ( x _ { j } ) \in R ^ {~c_{ T } \times h _ { T } \times w _ { T } }$, and the penultimate layer feature of the student network be represented as $f ^ { S } ( x _ { i } ) \in R ^ {~c_{S} \times h_{S} \times w_{ S } }$.
The student's feature is then passed through a projector to match the teacher's channel dimension, denoted as $S ^ { i } \in R ^ {~c_{T} \times h_{S} \times w_{ S } }$.
Here, $ c_{ T }$ and $c_{S}$ denote the number of feature channels for teacher and student networks, respectively, while $h _ { T },  w _ { T }$ and $h_{ S }, w_{ S }$ represent the spatial dimensions of teacher and student feature maps.
The classifier of the teacher network, denoted as $ fc ^ { T }$, first receives the global pooled representation of the feature map $ T ^ { i }$, denoted as $\bar { T } ^ { i } = G A P ( T ^ { i } ) \in R ^ {~c_{ T } \times 1 \times 1 }$, and then outputs the maximum softmax probability as $P^{i} = \max\limits_{c} \; \mathrm{Softmax}\!\left(fc^{T}(\bar{T}^{i})\right)_c$.
\begin{figure*}[t]
    \centering
    \includegraphics[width=0.95\textwidth]{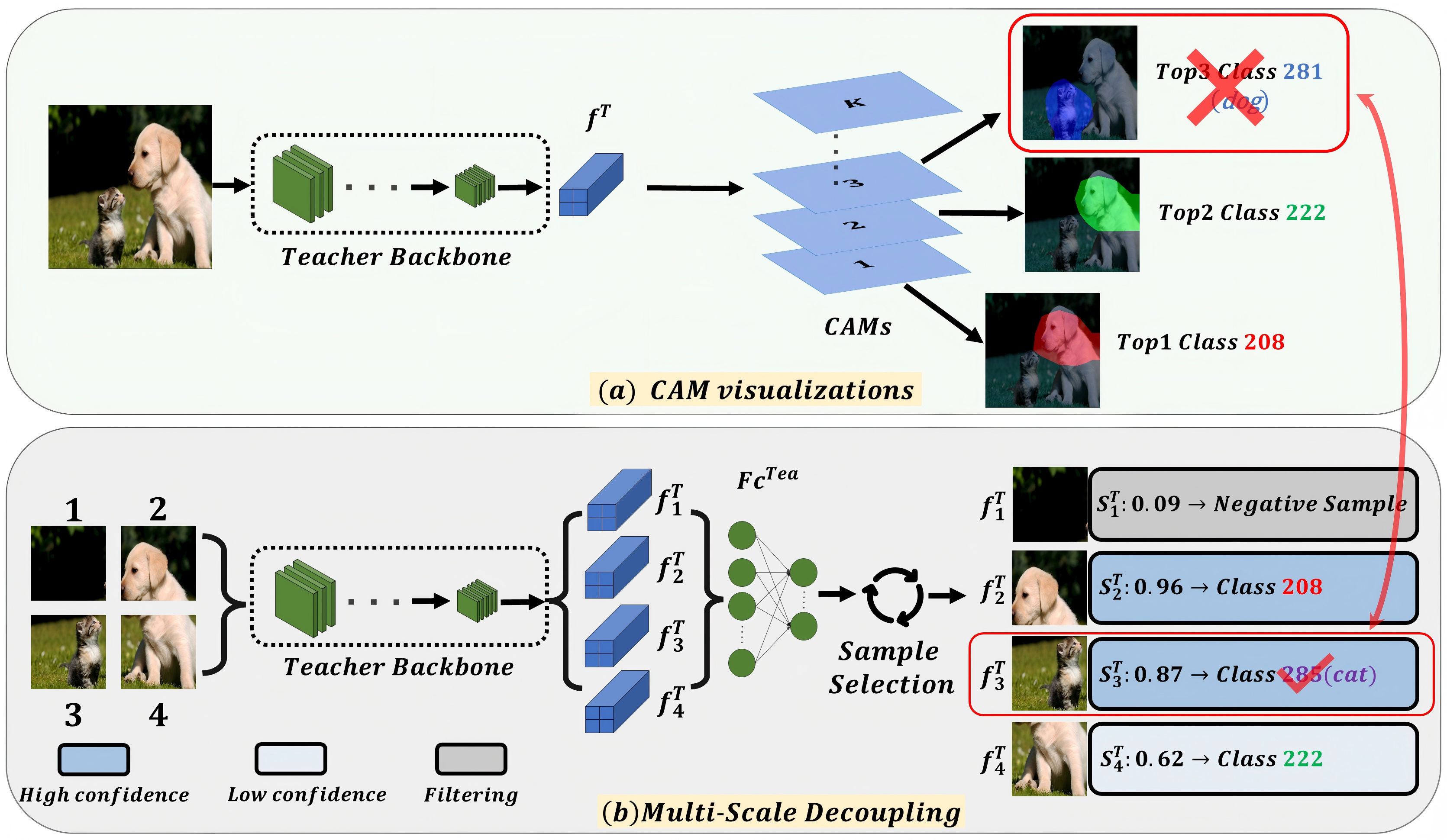}
\caption{
\textbf{Visualization of The Multi-Scale Decoupling Process.}
(a) CAM visualizations.
A single image is fed into the teacher network, and the top-3 class activation maps (CAMs) are visualized.
(b) Multi-scale Decoupling.
Multi-scale pooling is equivalently achieved by partitioning a single input into multiple local regions and  extracting their features separately, followed by sample selection to further process the resulting feature samples.
}
\vskip -0.2in
\label{Fig4}
\end{figure*}
\subsection{Multi-Scale Decoupling}
\label{MSD}
As illustrated in Figure~\ref{Fig4} (a), when directly transferring the global feature of the teacher network, CAM visualizations reveal that different local regions within an individual global feature attend to distinct high-dimensional semantics.
Due to the coupling among these local features, some local features are prone to semantic confusion, resulting in incorrect predictions. 
For example, in the red box of (a), the blue-masked region should correspond to the category \textit{cat}, but CAM highlights class 281 (\textit{dog}), which is clearly incorrect.
In contrast, in (b), after applying multi-scale feature decoupling, an individual global feature is decoupled into multiple local features and processed separately.
The red box in (b) shows that the prediction corresponds to class 285 (\textit{cat}), which is correct.
This comparison demonstrates that directly transferring global features fails to capture the rich fine-grained information contained within an individual global feature and is unable to effectively decouple the dependencies among different local regions.
Consequently, semantic confusion arises, hindering the student network from fully acquiring the teacher’s knowledge and ultimately resulting in suboptimal performance.

For the multi-scale decoupling designed in the feature space, the overall process can be divided into two steps.
The first step is \textbf{multi-scale pooling}, where this step employs sliding pooling windows of different scales to extract as much fine-grained knowledge as possible from an individual global feature.
The second step is \textbf{sample selection}, in which the obtained samples are analyzed and those containing only irrelevant information are filtered.
The remaining samples are subsequently divided into high-confidence and low-confidence groups based on their confidence scores, with each group being processed differently.

\textbf{Multi-Scale Pooling.}
To capture rich fine-grained knowledge, this work first extracts the feature maps $S ^ { i }$ and $ T ^ { j }$  from the penultimate layer of both the student and teacher networks, respectively.
Multi-scale sliding pooling window is then performed by setting different kernel sizes and strides.
This operation generates multiple pooled feature samples from various scales and spatial locations within an individual global feature.
These pooled samples are denoted as $ S _ { m } ^ { i } \in R ^ {~c _ { T } \times 1 \times 1 }$ and $T _ { n } ^ { j } \in R ^ {~c _ { T } \times 1 \times 1 }$, where $ m, n = 1, 2, ..., M$, and $M$ denote the total number of pooled samples (including the globally pooled feature obtained via $GAP$) derived from an individual global feature.
The condition $m=n$ indicates that the pooled regions share the same pooling window size and pooling position. 
When $i=j$,  the teacher and student networks are fed with the same input.
\begin{figure}[t] 
    \centering

    \subfigure[\label{ResNet-34 (w/o MSP)}]{
        \includegraphics[width=0.45\linewidth]{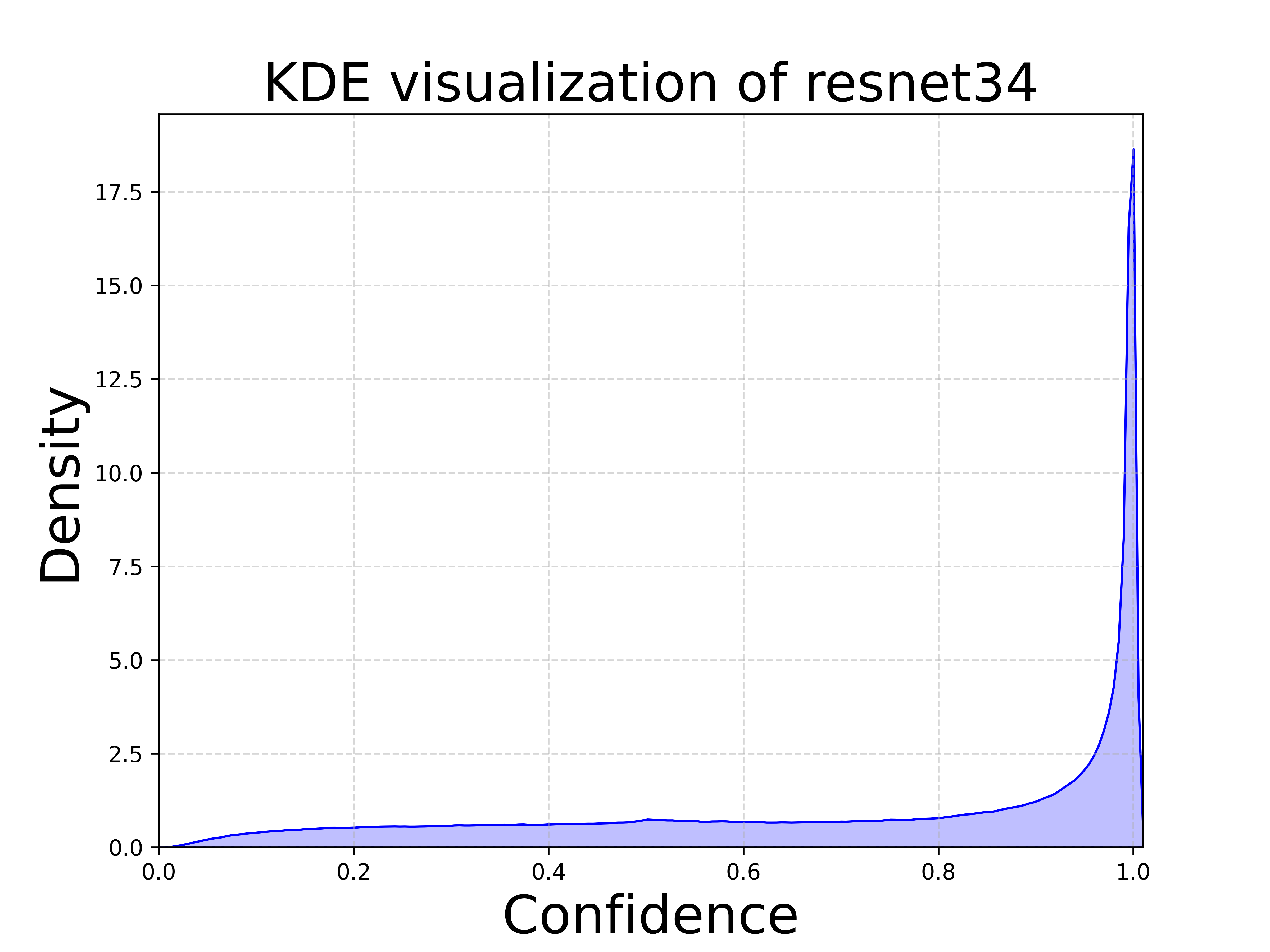}
    }
    \hfill
    \subfigure[\label{ResNet-34 (w/ MSP)}]{
        \includegraphics[width=0.45\linewidth]{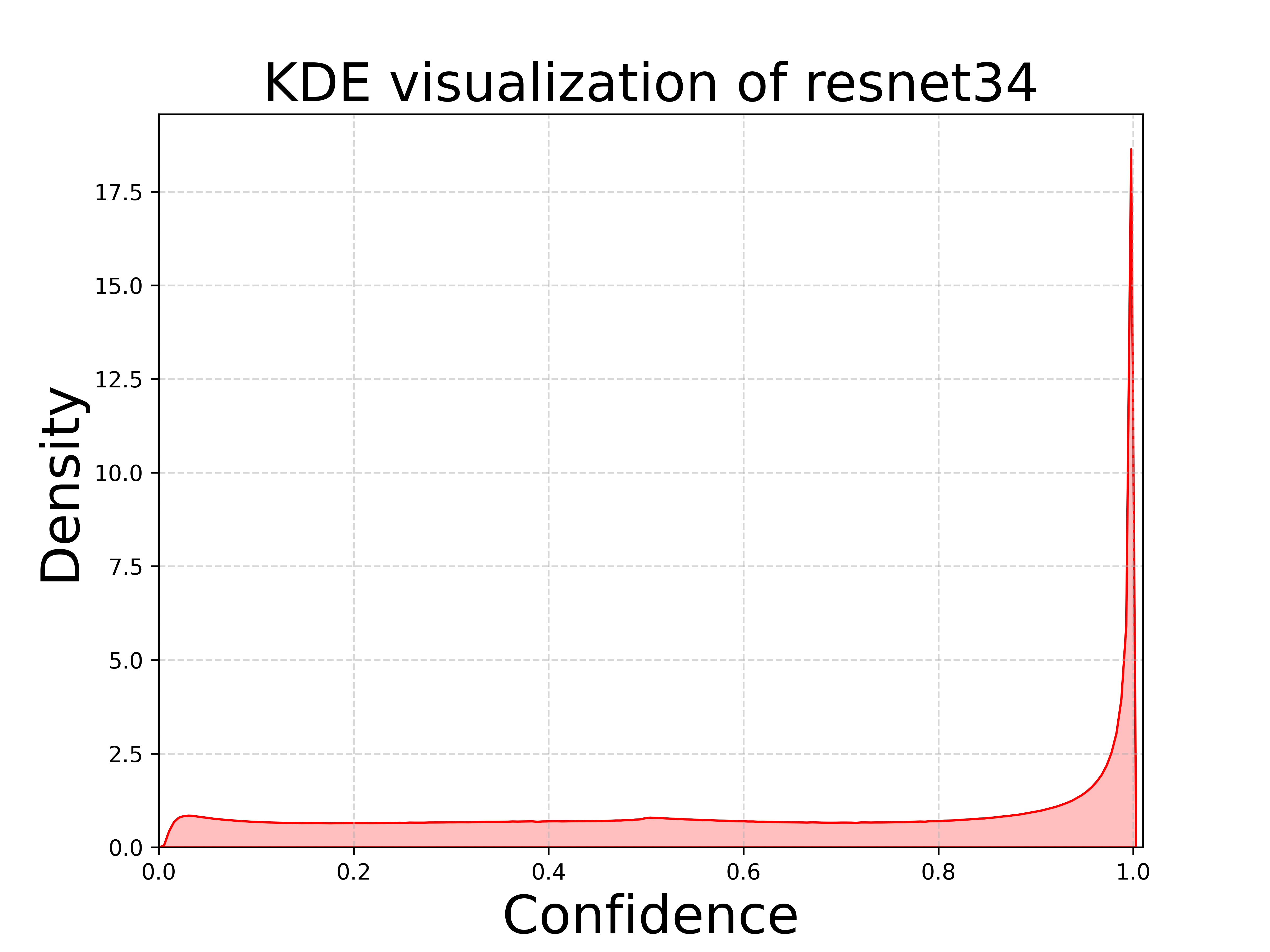}
    }

    \subfigure[\label{Swin-Tiny (w/o MSP)}]{
        \includegraphics[width=0.45\linewidth]{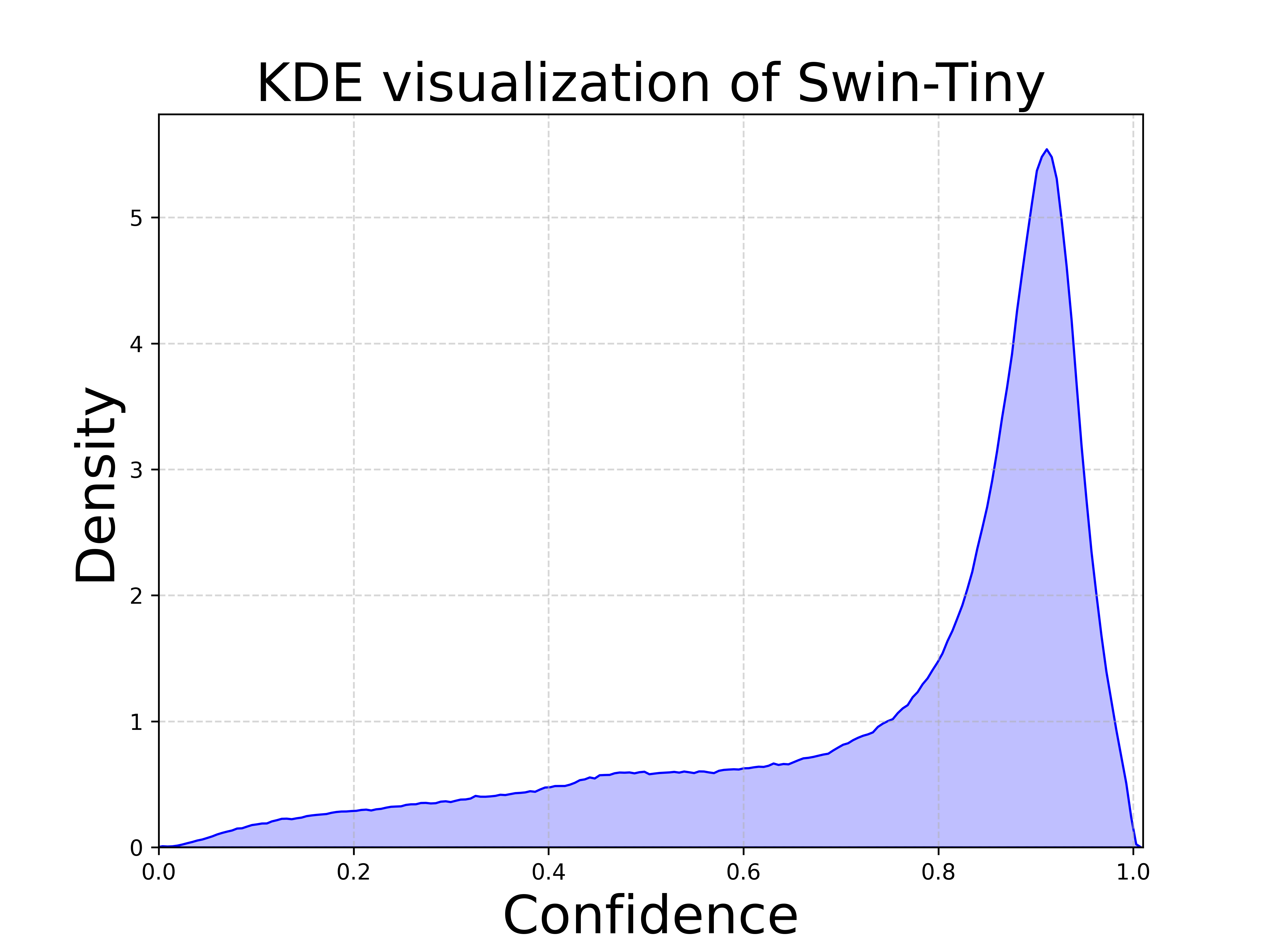}
    }
    \hfill
    \subfigure[\label{Swin-Tiny (w/ MSP)}]{
        \includegraphics[width=0.45\linewidth]{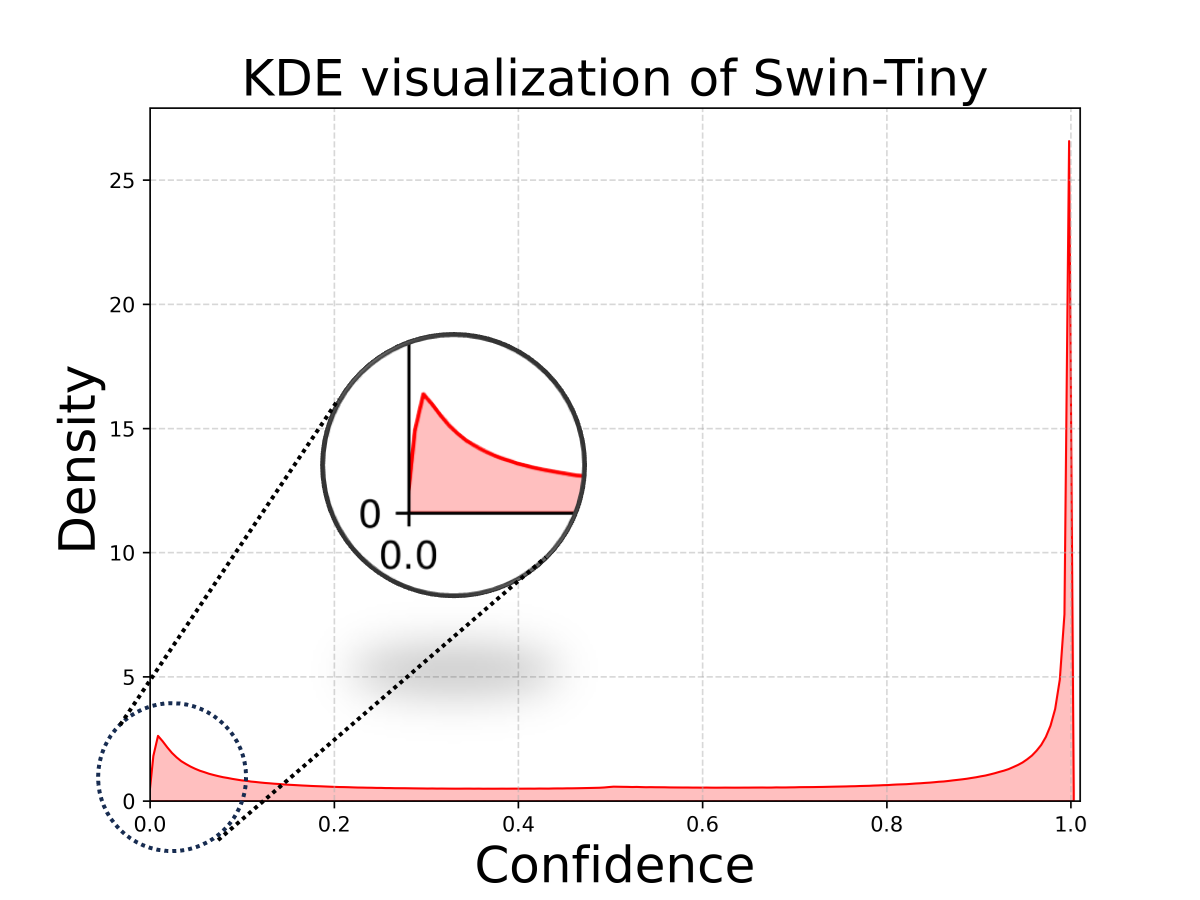}
    }

    \subfigure[\label{Mixer-B/16 (w/o MSP)}]{
        \includegraphics[width=0.45\linewidth]{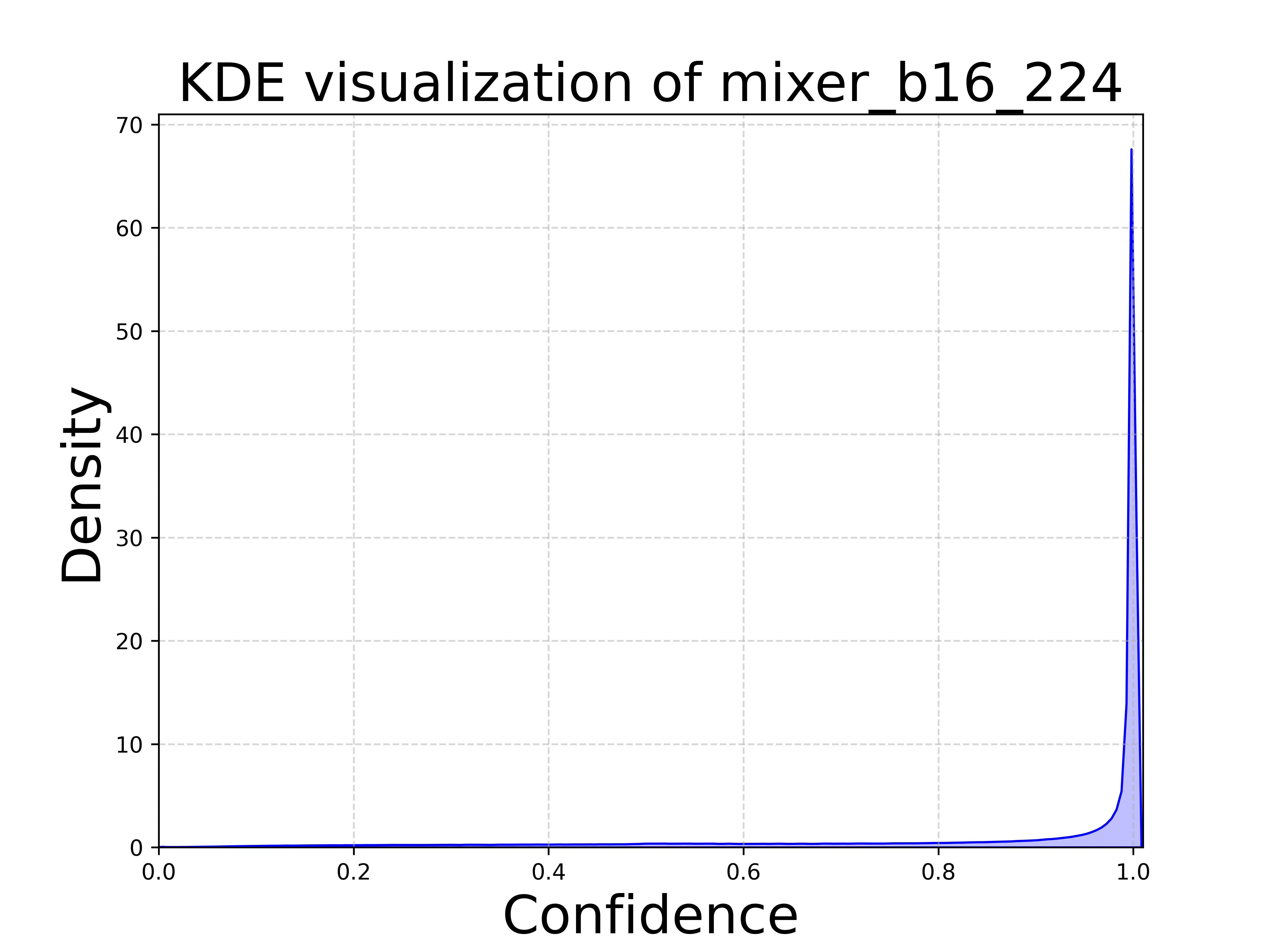}
    }
    \hfill
    \subfigure[\label{Mixer-B/16 (w/ MSP)}]{
        \includegraphics[width=0.45\linewidth]{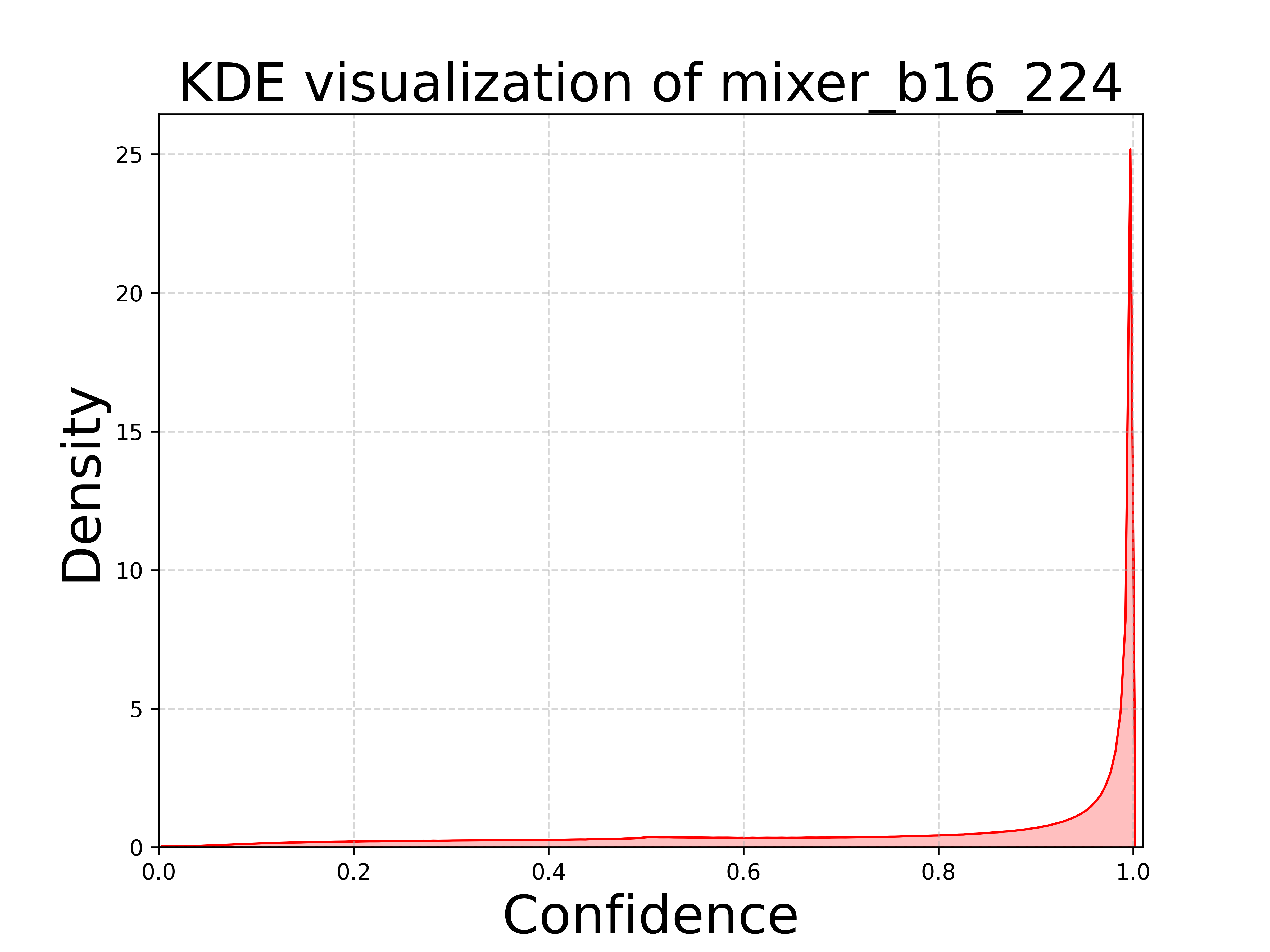}
    }

    \caption{\textbf{Maximum Softmax Probability Distributions on ImageNet with and without Multi-Scale Pooling (MSP).} The first column shows the maximum softmax probability distributions of samples from different teacher networks without MSP, while the second column shows the corresponding distributions with MSP.}
    \label{fig5}
\vskip -0.2in
\end{figure}

\textbf{Sample Selection.}
For the $j$-th input image $x_{ j }$ fed into the teacher network, the extracted feature map $T ^ { j }$ is processed through multi-scale pooling, resulting in $M$ local feature samples, denoted as $T _ { n } ^ { j }$ for $ n = 1, 2, ..., M$.
Each of these pooled feature maps is then fed into the pretrained classifier $ fc ^ { T }$  of the teacher network to obtain its maximum softmax probability, yielding $P_ { n }^{j} = \max\limits_{c} \; \mathrm{Softmax}\!\left(fc^{T}( T _ { n }^ { j })\right)_c$.
As illustrated in Figure~\ref{Fig4}(b), different regions within a single image attend to different categories of information.
After multi-scale pooling, it can be observed that some features capture only background, which corresponds to non-informative or irrelevant information.
For example, the local feature denoted as $f _ { 1 } ^ { T }$ in Figure~\ref{Fig4}(b) lacks any target category information and is therefore regarded exclusively as a negative sample.
The maximum softmax probabilities produced by such regions are consistently low and can be removed by applying a filtering threshold.
In contrast, as shown by the local regions denoted as $f _ { 2 } ^ { T }, f _ { 3 } ^ { T }, f _ { 4 } ^ { T }$ in Figure~\ref{Fig4}(b), the remaining samples after filtering exhibit different levels of discriminative power with respect to target category information.
Specifically, $f _ { 2 } ^ { T }$ and $ f _ { 3 } ^ { T }$  contain relatively stronger discriminative information, whereas $ f _ { 4 } ^ { T }$ provides comparatively weaker information.
This difference can be directly identified through the maximum softmax probabilities $S_ { n }^{j}$ associated with each local feature.
As shown in Figure~\ref{fig5}, the maximum softmax probabilities of feature samples are statistically analyzed both without multi-scale pooling ((a), (c), (e)) and with multi-scale pooling ((b), (d), (f)).
It can be observed that, after multi-scale pooling, the distribution of maximum softmax probabilities ((b), (d), (f)) exhibits a long-tailed pattern: the majority of pooled feature samples are concentrated within a narrow interval of relatively high probabilities, while only a small portion of samples are distributed across the remaining broader interval.
Furthermore, as shown in Figure~\ref{fig5} (d), a local peak emerges in the low-probability region, which can be attributed to the presence of a relatively larger number of non-informative feature samples introduced by multi-scale pooling.
Therefore, to discard non-informative samples while fully exploiting the knowledge contained in each sample of the teacher network, two thresholds, $\alpha$ and $\beta$, are introduced in the sample selection process.
First, the filtering threshold $\alpha$ is applied to eliminate samples that contain only non-informative information (i.e., $P_{n}^{j} < \alpha$).
Then, the remaining samples are divided into low-confidence and high-confidence groups based on a second threshold $\beta$ (i.e., $\alpha \leq P_{n}^{j} < \beta$ for low-confidence samples, and $P_{n}^{j} \geq \beta$ for high-confidence samples).
To address the imbalance between these two groups within each batch, an inverse-frequency weighting scheme is adopted.
Specifically, the weight of each low-confidence sample is set to half of the reciprocal of the number of high-confidence samples in the same batch 
($w_{low} = \tfrac{1}{2} \cdot \tfrac{1}{N_{high}}$),  while the weight of each high-confidence sample is set to half of the reciprocal of the number of low-confidence samples in the same batch 
($w_{high} = \tfrac{1}{2} \cdot \tfrac{1}{N_{low}}$).
The overall weighting scheme can be formulated as follows:
\begin{equation}
w_{sample}(S_n^j) =
\begin{cases}
0, & P_n^j < \alpha \\[8pt]
\tfrac{1}{2} \cdot \dfrac{1}{N_{\text{high}}}, & \alpha \leq P_n^j < \beta \\[12pt]
\tfrac{1}{2} \cdot \dfrac{1}{N_{\text{low}}}, & P_n^j \geq \beta
\end{cases}
\label{eq4}
\end{equation}
\subsection{Sample-Wise Contrastive Loss}
Compared with previous contrastive representation distillation methods~\cite{tian2019contrastive}, MSDCRD eliminates the reliance on a large memory bank that stores feature representations from teacher and student networks.
Instead, it relies solely on a single batch of inputs and, by employing the proposed multi-scale feature decoupling, generates a diverse collection of feature samples with rich and varied semantic information.
These samples are obtained not only from different input images but also from distinct local regions within the same image.

For a batch of input data with size $B$, $B$ feature maps are obtained from the student and teacher networks, denoted as $S^{i}$ and $T^{j}$, respectively.
After applying multi-scale decoupling, each feature map yields $M$ pooled local features, resulting in $N = M \times B$ decoupled feature samples from each network, denoted as $S_{m}^{i}$ and $T_{n}^{j}$.
When $i = j$ and $m = n$, the pair consisting of the student feature sample $S _ { m }^ { i }$ and the teacher feature sample $T _ { m }^ { i }$, which correspond to the same input sample and the same pooled local region, is defined as a \textbf{positive pair}.
Conversely, when $i \neq j$ or $m \neq n$, the pair consisting of the student feature sample $S _ { m }^ { i }$ and the teacher feature sample $T _ { n }^ { j }$, which correspond to different input images or different pooled local regions, is defined as a \textbf{negative pair}.
The objective is to push closer the representations $S _ { m }^ { i }$ and $T _ { m }^ { i }$, while pushing apart $S _ { m }^ { i }$ and $T _ { n }^ { j }$ (where $i \neq j$ or $m \neq n$).
Specifically, this work considers the joint distribution $p(S _ { m }^ { i }, T _ { n }^ { j })$ between the student network feature samples and the teacher network feature samples, as well as the product of the marginal distributions $p(S _ { m }^ { i })p(T _ { n }^ { j })$, and ultimately aim to maximize the mutual information between positive pairs
\begin{equation}
I ( S _ { m } ^ { i } , T _ { m } ^ { i } ) = E _ { p ( S _ { m } ^ { i } , T _ { m } ^ { i } ) } ( \log ( \frac { p ( S _ { m } ^ { i } , T _ { m } ^ { i } ) } { p ( S _ { m } ^ { i } ) p ( T _ { m } ^ { i } ) } ) )
\label{eq5}
\end{equation}
where $E _ { p ( S _ { m } ^ { i } , T _ { m } ^ { i } ) } (\cdot)$ indicates the expectation over positive pairs, $p ( S _ { m } ^ { i } , T _ { m } ^ { i } )$ denotes the joint distribution of positive pairs.

To achieve this goal, a distribution $q$ is defined and a latent variable $V$ is introduced to determine whether, for a given pair of student–teacher feature samples $(S _ { m }^ { i }, T _ { n }^ { j })$ , the distribution $q$ is drawn from the joint distribution $p ( S _ { m } ^ { i } , T _ { n } ^ { j } )$ or from the product of the marginal distributions $p(S _ { m }^ { i })p(T _ { n }^ { j })$
\begin{equation}
q ( S _ { m } ^ { i } , T _ { n } ^ { j } | V = 1 ) = p ( S _ { m } ^ { i } , T _ { n } ^ { j } )
\label{eq6}
\end{equation}
\begin{equation}
q ( S _ { m } ^ { i } , T _ { n } ^ { j } | V = 0 ) = p ( S _ { m } ^ { i } ) p ( T _ { n } ^ { j } )
\label{eq7}
\end{equation}

Building on the definition above, the latent variable $V$  is further specified using the decoupled feature samples obtained in Section~\ref{MSD}.
For positive pairs, where the student and teacher features originate from the same input and the same pooled local region, the samples are considered correlated and set to $V = 1$.
For negative pairs, derived from different inputs or different pooled regions, the decoupled features are regarded as uncorrelated and set to $V = 0$.

The prior probability of the latent variable $V$ can be easily derived as
\begin{equation}
q ( V = 1 ) = \frac { 1 } { N } , q ( V = 0 ) = \frac { N - 1 } { N }
\label{eq8}
\end{equation}

According to Bayes' theorem, the posterior formula for the variable $V$ can be written as
\begin{align}
\small
&q ( V = 1 | S_{m}^{i}, T_{n}^{j} )\nonumber\\
&= \frac { q ( S _ { m } ^ { i } , T _ { n } ^ { j } | V = 1 ) q ( V = 1 )} { q ( S _ { m } ^ { i } , T _ { n } ^ { j } | V = 1 ) q ( V = 1 ) + q ( S _ { m } ^ { i } , T _ { n } ^ { j } | V = 0 ) q ( V = 0 )}\nonumber\\
&= \frac {p ( S _ { m } ^ { i } , T _ { n } ^ { j } )} {p ( S _ { m } ^ { i } , T _ { n } ^ { j } ) + (N-1)p ( S _ { m } ^ { i } ) p ( T _ { n } ^ { j } )}
\label{eq9}    
\end{align}

By taking the negative logarithm of both sides of Equation~\ref{eq9} and exploiting the monotonicity of the logarithmic function, the following expression is obtained
\begin{align}
\small
&- \log q ( V = 1 | S _ { m } ^ { i } , T _ { n } ^ { j } )\nonumber\\
& = \log ( 1 + \frac { ( N - 1 ) p ( S _ { m } ^ { i } ) p ( T _ { n } ^ { j } ) } { p ( S _ { m } ^ { i } , T _ { n } ^ { j } ) } )\nonumber\\
& \geq \log ( N - 1 ) - \log ( \frac { p ( S _ { m } ^ { i } , T _ { n } ^ { j } ) } { p ( S _ { m } ^ { i } ) p ( T _ { n } ^ { j } ) } )
\label{eq10}    
\end{align}

To maximize the mutual information between positive pairs, the expectation of both sides of Equation~\ref{eq10} is taken w.r.t. $q ( S _ { m } ^ { i }, T _ { n } ^ { j } | V = 1 )$ (which is equivalent to $p ( S _ { m } ^ { i }, T _ { n } ^ { j } )$, where $i = j$ and $m = n$).
After rearrangement, the form of Equation~\ref{eq5} is obtained
\begin{align}
\small
&I ( S _ { m } ^ { i } , T _ { m } ^ { i } )\nonumber\\
&\geq \log ( N - 1 ) + E _ { p ( S _ { m } ^ { i } , T _ { m } ^ { i } )} (\log q ( V = 1 | S _ { m } ^ { i } , T _ { n } ^ { j } ))
\label{eq11}
\end{align}

After removing the constant term on the right-hand side of Equation~\ref{eq11}, the lower bound of the mutual information between positive pairs is obtained, denoted as $MI\ bound$.
\begin{equation}
\small
 MI\ bound = E _ { p ( S _ { m } ^ { i } , T _ { m } ^ { i } )} (\log q ( V = 1 | S _ { m } ^ { i } , T _ { n } ^ { j } ))
\label{eq12}
\end{equation}

Maximizing the lower bound $MI\ bound$ is equivalent to maximizing $I ( S _ { m } ^ { i } , T _ { m } ^ { i } )$.
However, since the true distribution $q ( V = 1 | S _ { m } ^ { i } , T _ { n } ^ { j } )$ is difficult to obtain, an approximate distribution $ \tilde { q } ( V = 1 | S _ { m } ^ { i } , T _ { n } ^ { j } )$ is constructed, given by
\begin{equation}
\small
\tilde { q } ( V = 1 | S _ { m } ^ { i } , T _ { n } ^ { j } ) = \frac { e x p ( sim ( S _ { m } ^ { i } , T _ { m } ^ { i } ) ) } { \sum _ { j = 1 } ^ { B } \sum _ { n = 1 } ^ { M } e x p ( sim ( S _ { m } ^ { i } , T _ { n } ^ { j } ) ) }
\label{eq13}
\end{equation}
where $sim(\cdot)$ is the cosine similarity function between two feature samples.
Therefore, when maximizing Equation~\ref{eq13}, it is equivalent to maximizing the mutual information between positive pairs
\begin{equation}
\max E _ { p ( S _ { m } ^ { i } , T _ { m } ^ { i } ) } ( \log \tilde { q } ( V = 1 | S _ { m } ^ { i } , T _ { n } ^ { j } ) )
\label{eq14}
\end{equation}

Before computing the loss, each teacher and student feature is individually normalized using L2 normalization.
At the same time, to filter out non-informative samples and ensure sufficient learning from each sample, the inverse-frequency weighting scheme introduced in Section~\ref{MSD} is incorporated into this loss function. 
The resulting sample-wise contrastive loss is defined as
\begin{equation}
\small
\begin{split}
&\mathcal{L}_{\text{sample}} =\\
& - \frac{1}{N_{\text{high}} + N_{\text{low}}}
\sum_{i=1}^{B} \sum_{m=1}^{M} 
w(P_{m}^{i}) \cdot 
\log \frac{\exp \big( \mathrm{sim}(\hat{S}_{m}^{i}-\bar{S}, \hat{T}_{m}^{i}-\bar{T}) \big)}
{Z_{n}^{j}}
\end{split}
\label{eq15}
\end{equation}
where $N_{\text{high}}$ is the number of high-confidence samples, and $N_{\text{low}}$ denotes the number of low-confidence samples.
$\hat{S}_{m}^{i}$ and $\hat{T}_{m}^{i}$ represent the student and teacher features after applying L2 normalization.
$\bar{S}$ and $\bar{T}$ denote the mean features of the student and teacher within a single batch.
$sim(\cdot)$ computes the cosine similarity between two feature samples, and $Z_{n}^{j}$ is defined in Equation~\ref{eq16}.
\begin{equation}
Z_{n}^{j} = \sum_{j=1}^{B} \sum_{n=1}^{M} 
\exp \big( \mathrm{sim}(\hat{S}_{m}^{i}-\bar{S}, \hat{T}_{n}^{j}-\bar{T}) \big)
\label{eq16}
\end{equation}
\subsection{Feature-Wise Contrastive Loss}
While the sample-wise contrastive loss enables the transfer of sample-level relational knowledge from the teacher network to the student network, it only captures sample-wise alignment.
In practice, however, the relationships among feature-wise themselves are also highly informative and valuable.
For instance, within an output feature of a network, different elements may capture distinct semantic directions, such as encoding whether an object belongs to the category of animals.
Consider three input samples: a cat, a dog, and a car. 
When processed by the network, the features of cat and dog exhibit stronger correlation along the “animal-related” semantic direction compared to the cat and car pair, as cats and dogs both belong to the category of animals in the real world.
Transferring such feature-wise relational knowledge to the student network is therefore crucial. 
Moreover, even among samples of the same category, the intrinsic intra-class variance of semantic similarities is also informative, as it reveals which feature samples are more discriminative within a category.

Therefore, this work introduces a feature-wise contrastive loss. 
First, the following threshold-based filtering strategy is defined
\begin{equation}
w_{\text{feature}}(P_{m}^{i}) =
\begin{cases}
0, & P_{m}^{i} < \alpha \\[6pt]
1, & P_{m}^{i} \geq \alpha
\end{cases}
\label{eq17}
\end{equation}
where $\alpha$ is the filtering threshold.

By applying this strategy, student and teacher samples that contain no informative feature content are discarded.
The remaining student and teacher feature samples, after transposition, are denoted as
\begin{equation}
\tilde{S}_{ch}^{\top} = \big(w_{\text{feature}}(P_{m}^{i})\cdot\hat{S}_{m}^{i} \big)^{\top},
\quad
\tilde{T}_{ch}^{\top} = \big(w_{\text{feature}}(P_{m}^{i})\cdot\hat{T}_{m}^{i} \big)^{\top}
\label{eq18}
\end{equation}
while $ch$ denotes the channel index of the feature with $ch \in [0, 1, 2, \dots, C-1]$ ,the means of the transposed filtered student and teacher feature samples are denoted as $\bar{\tilde{S}}^{\top}$ and $\bar{\tilde{T}}^{\top}$, respectively.
The resulting sample-wise contrastive loss is defined as
\begin{equation}
\small
\begin{split}
&\mathcal{L}_{\text{feature}} =\\
& - \frac{1}{C}\sum_{ch=1}^{C}  
\log \frac{\exp \big( \mathrm{sim}(\tilde{S}_{ch}^{\top} - \bar{\tilde{S}}^{\top}, \tilde{T}_{ch}^{\top} - \bar{\tilde{T}}^{\top} ) \big)}
{F_{j}}
\end{split}
\label{eq19}
\end{equation}
where $C$ denotes the number of channels in the teacher features, and $F_{n}^{j}$ is defined in Equation~\ref{eq20}.
\begin{equation}
F_{j} = \sum_{j=1}^{C} 
\exp \big( \mathrm{sim}(\tilde{S}_{ch}^{\top} - \bar{\tilde{S}}^{\top},\tilde{T}_{j}^{\top} - \bar{\tilde{T}}^{\top}) \big)
\label{eq20}
\end{equation}

As shown in Algorithm~\ref{algorithm1} , the overall training loss $L_{tr}$ consists of the target loss $\mathcal{L}_{obj}$, the sample-wise contrastive loss $\mathcal{L}_{sample}$, 
and the feature-wise contrastive loss $\mathcal{L}_{feature}$, i.e.,
\begin{equation}
\mathcal{L}_{tr} = \mathcal{L}_{obj} 
+ \lambda_{1}\mathcal{L}_{sample} 
+ \lambda_{2}\mathcal{L}_{feature}
\label{eq21}
\end{equation}
where $\lambda_{1}$ and $\lambda_{2}$ are trade-off hyperparameters that balance the contributions of the sample-wise and feature-wise contrastive losses.

\begin{algorithm}[H]
\caption{MSDCRD Loss}
\label{alg:msdcrd}
\begin{algorithmic}[1]
\STATE {\textsc{Input}} : $\mathcal{X}, \mathcal{Y}$, input images, input labels\\
            \hspace*{3.5em} $T, S$, teacher backbone, student backbone\\
            \hspace*{3.5em} $\lambda_{1}, \lambda_{2}$, hyperparameters
\STATE {\textsc{Output}} : $\mathcal{L}_{tr}$
\FOR{each mini-batch $(\mathcal{X}, \mathcal{Y})$}
    \STATE $F_T \gets T(\mathcal{X}), \quad F_S \gets S(\mathcal{X})$
    \STATE $T_{m}^{i}, S_{m}^{i} \gets \text{MSP}(F_T, F_S)$
    \STATE $P_{m}^{i} = \max_{c} \;\mathrm{Softmax}(fc^{T}(T_{m}^{i}))_{c}$
    \STATE $w_{\text{sample}}(P_{m}^{i}), w_{\text{feature}}(P_{m}^{i}) \gets \text{SS}(P_{m}^{i})$
    \STATE $\mathcal{L}_{sample} \gets T_{m}^{i},S_{m}^{i}, w_{\text{sample}}(P_{m}^{i})$
    \STATE $\mathcal{L}_{feature} \gets T_{m}^{i}, S_{m}^{i}, w_{\text{feature}}(P_{m}^{i})$
    \STATE $\mathcal{L}_{obj} \gets F_S, Y$
    \STATE $\mathcal{L}_{tr} = \mathcal{L}_{target} 
    + \lambda_{1}\mathcal{L}_{sample} 
    + \lambda_{2}\mathcal{L}_{feature}$
\ENDFOR
\STATE \textbf{return} $\mathcal{L}_{tr}$
\end{algorithmic}
\label{algorithm1}
\end{algorithm}
\begin{table*}[ht]
\vskip -0.2in
\renewcommand{\arraystretch}{1.5}
\centering
\caption{\textbf{Comparison of Homogeneous Distillation Results on CIFAR-100.}  The best and second-best results are emphasized in \textbf{bold} and \underline{underlined} cases. Results where MSDCRD outperforms the teacher network are marked with \textbf{*}.}
\scalebox{0.9}{
\begin{tabular}{cc|cc|c|cc|ccccc:cc}
\hline
\multirow{2}{*}{Teacher} & \multirow{2}{*}{Student} & \multicolumn{2}{|c|}{From Scratch} & \multicolumn{1}{c|}{Logits-based} & \multicolumn{2}{c|}{Attention-based} & \multicolumn{5}{c:}{Feature-based} & \multicolumn{2}{c}{Feature-based}\\
\cline{3-14} 
        &             & T. & S. & KD  & AT & CAT-KD & FitNet & RKD & SimKD & NORM & ReviewKD & CRD & \textbf{MSDCRD} \\
\hline
ResNet110 & ResNet32 & 74.31 & 71.14 & 73.08 & 72.31 & 73.62 & 71.06 & 71.82 & \underline{73.92} & 73.67 & 73.89 & 73.48& \cellcolor{gray!20}\textbf{*74.61}($\uparrow$+1.13)\\
ResNet50 & MobileNetV2 & 79.34 & 64.60 & 67.35 & 58.58 & \underline{71.36} & 63.16 & 64.43 & 69.97 & 70.56 & 69.89 & 69.11 & \cellcolor{gray!20}\textbf{71.51}($\uparrow$+2.40)\\
WRN-40-2 & WRN-16-2  & 75.61 & 73.26 & 74.92 & 74.08 & 75.60 & 73.58 & 73.35 & 75.53 & 76.62 & \underline{76.12} & 75.48& \cellcolor{gray!20}\textbf{*76.66}($\uparrow$+1.18)\\
WRN-40-2 & ResNet8$\times$4  & 75.61 & 72.50 & 73.97 & 74.11 & \underline{75.38} & 74.61 & 75.26 & 75.29 & N/A & 74.34 & 75.24 & \cellcolor{gray!20}\textbf{*76.93}($\uparrow$+1.69)\\
ResNet32$\times$4 & ResNet8$\times$4  & 79.42 & 72.50 & 73.33 & 73.44 & 76.91 & 73.50 & 71.90 & \textbf{78.08} & 76.98 & 75.63 & 75.51& \cellcolor{gray!20}\underline{77.43}($\uparrow$+1.92)\\
ResNet32$\times$4 & WRN-40-2 & 79.42 & 75.61 & 77.70 & 77.43 & 78.59 & 77.69 & 77.82 & \underline{79.29} & N/A & 78.96 & 78.15 & \cellcolor{gray!20}\textbf{*80.34}($\uparrow$+2.19)\\
ResNet32$\times$4 & ShuffleNetV1  & 79.42 & 70.50 & 74.07 & 71.73 & \underline{78.26} & 73.59 & 72.28 & 77.18 & 77.42 & 77.45 & 75.11 & \cellcolor{gray!20}\textbf{78.65}($\uparrow$+3.54)\\
VGG13 & VGG8 & 74.64 & 70.36 & 72.98 & 71.43 & 74.65 & 71.02 & 71.48 & \underline{74.89} & 74.46 & 74.84 & 73.94 & \cellcolor{gray!20}\textbf{*75.65}($\uparrow$+1.71)\\
\hline
\end{tabular}
}
\label{table1}
\end{table*}
\begin{table*}[ht]
\renewcommand{\arraystretch}{1.5}
\centering
\caption{\textbf{Comparison of Homogeneous Distillation Results on ImageNet.} The best and second-best results are emphasized in \textbf{bold} and \underline{underlined} cases. Results where MSDCRD outperforms the teacher network are marked with \textbf{*}.}
\scalebox{0.93}{
\begin{tabular}{cc|cc|ccc|cc|cccc:cc}
\hline
\multirow{2}{*}{Teacher} & \multirow{2}{*}{Student} & \multicolumn{2}{c|}{From Scratch} & \multicolumn{3}{c|}{Logits-based} & \multicolumn{2}{c|}{Attention-based} & \multicolumn{4}{c:}{Feature-based} & \multicolumn{2}{c}{Feature-based}\\
\cline{3-15} 
        &            & T. & S. & KD & OFA & DIST & AT & CAT-KD & RKD & SimKD & NORM & ReviewKD & CRD & \textbf{MSDCRD} \\
\hline
ResNet34 & ResNet18 & 73.31 & 69.75 & 70.68 & 72.10 & 71.89 & 70.59 & 71.26  & 71.34 & 71.66 & \underline{72.14} & 71.61 & 71.17 & \cellcolor{gray!20}\textbf{72.21}($\uparrow$+1.04)\\
\hline
\end{tabular}
}
\vskip -0.1in
\label{table2}
\end{table*}
\section{Experiments}
This work conducts experiments across multiple tasks.
First, classification experiments are conducted on multiple datasets under homogeneous model settings.
Meanwhile, as demonstrated in Section~\ref{FRG}, feature discrepancies are more pronounced across heterogeneous models, making knowledge transfer across heterogeneous models more challenging than in homogeneous cases.
To demonstrate the model-agnostic property of MSDCRD, classification experiments are further conducted across heterogeneous models on diverse vision datasets.
Moreover, to verify its generality, the proposed method is also evaluated on additional vision tasks. 
The detailed experimental settings are provided below.
\subsection{Classification}
\textbf{Datasets.}
Experiments are conducted on CIFAR-100~\cite{krizhevsky2009learning} and ImageNet~\cite{russakovsky2015imagenet}.
CIFAR-100 is a widely used image classification benchmark, containing 100 categories with 50,000 training images and 10,000 validation images, each with a resolution of 32×32. 
ImageNet is a larger-scale image classification dataset, consisting of images from 1,000 categories. 
The training set contains 1.28 million images, while the validation set includes 50,000 images.

\textbf{Implementation Details.}
Both homogeneous and heterogeneous knowledge distillation experiments are conducted on CIFAR-100 and ImageNet. The implementation details are summarized as follows.

On CIFAR-100, homogeneous distillation experiments are performed using commonly adopted CNN architectures in visual tasks. 
The proposed method is compared with several representative feature-based distillation approaches across widely used backbones, including VGG~\cite{simonyan2014very}, ResNet~\cite{he2016deep}, WideResNet~\cite{zagoruyko2016wide}, MobileNet~\cite{sandler2018mobilenetv2}, and ShuffleNet~\cite{ma2018shufflenet,zhang2018shufflenet}.
The training configuration follows the setting of~\cite{tian2019contrastive}. 
All models are trained for 240 epochs, with the learning rate decayed by a factor of 0.1 every 30 epochs after the initial 150 epochs. 
The batch size is fixed at 64 for each model. 
For heterogeneous distillation experiments, three representative architectures in vision tasks are considered: CNNs~\cite{liu2022convnet,sandler2018mobilenetv2,he2016deep}, Transformers~\cite{liu2021swin,dosovitskiy2020image}, and MLPs~\cite{tolstikhin2021mlp,touvron2022resmlp}.
Since ViTs and MLPs accept image patches as input, CIFAR-100 images are upsampled to a resolution of 224×224 for all heterogeneous experiments.
All models are trained for 300 epochs. CNN-based students are optimized using SGD, while Transformer-based and MLP-based students are trained using the AdamW optimizer. 
All the aforementioned experiments on both homogeneous and heterogeneous models based on the CIFAR-100 dataset are conducted on a single NVIDIA 3090 GPU.

On ImageNet, homogeneous distillation experiments are conducted using popular CNN architectures~\cite{he2016deep}, with comparisons made against recent feature-based distillation approaches. 
For heterogeneous distillation experiments, the same three representative architectures are adopted: CNNs~\cite{liu2022convnet,sandler2018mobilenetv2,he2016deep}, Transformers~\cite{liu2021swin}, and MLPs~\cite{tolstikhin2021mlp,touvron2022resmlp}.
In homogeneous settings, all models are trained for 100 epochs with an initial learning rate of 0.1, decayed by a factor of 0.1 every 30 epochs.
Under the heterogeneous settings, CNN-based students are trained for 100 epochs using the SGD optimizer, while Transformer-based and MLP-based students are trained for 300 epochs using the AdamW optimizer.
All homogeneous and heterogeneous experiments on ImageNet are conducted using four NVIDIA 3090 GPUs.

Each experiment is repeated three times, and the average top-1 accuracy is reported.

\begin{table*}[ht]
\vskip -0.2in
\renewcommand{\arraystretch}{1.2}
\centering
\caption{\textbf{Comparison of Heterogeneous Distillation Results on CIFAR-100.}  The best and second-best results are emphasized in \textbf{bold} and \underline{underlined} cases.}
\scalebox{1.05}{
\begin{tabular}{cc|cc|cccc|ccc:cc}
\hline
\multirow{2}{*}{Teacher} & \multirow{2}{*}{Student} &\multicolumn{2}{c|}{From Scratch} & \multicolumn{4}{c|}{Logits-based} & \multicolumn{3}{c:}{Feature-based} & \multicolumn{2}{c}{Feature-based}\\
\cline{3-13} 
        &       & T. & S. & KD  & DKD & DIST & OFAKD & FitNet & CC & RKD & CRD & \textbf{MSDCRD}\\
\hline
Swin-T & ResNet18 & 89.26 & 74.01 & 78.74 & 80.26 & 77.75 & \underline{80.54} & 78.87 & 74.19 &  74.11 & 77.63 & \cellcolor{gray!20}\textbf{84.47}($\uparrow$+6.79) \\
Mixer-B/16 & ResNet18 & 87.29 & 74.01 & 77.79 & 78.67 & 76.36 & \underline{79.39} & 77.15 & 74.26 & 73.75 & 76.42 & \cellcolor{gray!20}\textbf{79.51}($\uparrow$+3.09) \\
Swin-T & MobileNetV2 & 89.26 & 73.68 & 74.68 & 71.07 & 72.89 & \underline{80.98} & 74.28 & 71.19 &  69.00 & 79.80 & \cellcolor{gray!20}\textbf{82.74}($\uparrow$+2.94) \\
Mixer-B/16 & MobileNetV2 & 87.29 & 73.68 & 73.33 & 70.20 & 73.26 & \underline{78.78} & 73.78 & 70.73 & 68.95 & 78.15 & \cellcolor{gray!20}\textbf{78.86}($\uparrow$+0.71) \\
\hline
ConvNeXt-T &  DeiT-T  & 88.41 & 68.00 & 72.99 & 74.60 & 73.55 & \underline{75.76} & 60.78 & 68.01 & 69.79 & 65.94 & \cellcolor{gray!20}\textbf{85.52}($\uparrow$+19.58) \\
ConvNeXt-T &  Swin-P  & 88.41 & 72.63 & 76.44 & 76.80 & 76.41 & \underline{78.32} & 24.06 & 72.63 & 71.73 & 67.09 & \cellcolor{gray!20}\textbf{84.3}($\uparrow$+17.21) \\
Mixer-B/16 &  DeiT-T  & 87.29 & 68.00 & 71.36 & 73.44 & 71.67 & \underline{73.90} & 71.05 & 68.13 & 69.89 & 65.35 & \cellcolor{gray!20}\textbf{81.18}($\uparrow$+15.83) \\
Mixer-B/16 &  Swin-P  & 87.29 & 72.63 & 75.93 & 76.39 & 75.85 & \underline{78.93} & 75.20 & 73.32 & 70.82 & 67.03 & \cellcolor{gray!20}\textbf{80.23}($\uparrow$+13.20) \\
\hline
ConvNeXt-T &  ResMLP-S12  & 88.41 & 66.56 & 72.25 & 73.22 & 71.93 & \underline{81.22} & 45.47 & 67.70 & 65.82 & 63.35  & \cellcolor{gray!20}\textbf{86.87}($\uparrow$+23.52) \\
Swin-T &  ResMLP-S12  & 87.29 & 66.56 & 71.89 & 72.82 & 11.05 & \underline{80.63} & 63.12 & 68.37 & 64.66 & 61.72 & \cellcolor{gray!20}\textbf{86.45}($\uparrow$+24.41) \\
\hline
\end{tabular}
}
\label{table3}
\end{table*}
\begin{table*}[h]
\renewcommand{\arraystretch}{1.2}
\centering
\caption{\textbf{Comparison of Heterogeneous Distillation Results on ImageNet.} The best and second-best results are emphasized in \textbf{bold} and \underline{underlined} cases.}
\scalebox{1.05}{
\begin{tabular}{cc|cc|cccc|ccc:cc}
\hline
\multirow{2}{*}{Teacher} & \multirow{2}{*}{Student} &\multicolumn{2}{|c|}{From Scratch} & \multicolumn{4}{c|}{Logits-based} & \multicolumn{3}{c:}{Feature-based} & \multicolumn{2}{c}{Feature-based}\\
\cline{3-13} 
        &       & T. & S. & KD  & DKD & DIST & OFAKD & FitNet & CC & RKD & CRD & \textbf{MSDCRD}\\
\hline
Swin-T & ResNet18 & 81.38 & 69.75 & 71.14 & 71.10 & 70.91 & \underline{71.85} & 71.18 & 70.07 & 69.47 & 69.25 & \cellcolor{gray!20}\textbf{71.94}($\uparrow$+2.69) \\
Mixer-B/16 & ResNet18 & 76.62 & 69.75 & 70.89 & 69.89 & 70.66 & \textbf{71.38} & 70.78 & 70.05 & 69.46 & 68.40 & \cellcolor{gray!20}\underline{71.24}($\uparrow$+2.84) \\
Swin-T & MobileNetV2 & 81.38 & 68.87 & 72.05 & 71.71 & 71.76 & \underline{72.32} & 71.75 & 70.69 & 67.52 & 69.58 & \cellcolor{gray!20}\textbf{72.35}($\uparrow$+2.77) \\
Mixer-B/16 & MobileNetV2 & 76.62 & 68.87 & \underline{71.92} & 70.93 & 71.74 & \textbf{72.12} & 71.59 & 70.79 & 69.86 & 68.89 & \cellcolor{gray!20}71.34($\uparrow$+2.45) \\
\hline
ConvNeXt-T & Swin-N  & 82.05 & 75.53 & 77.15 & 77.00 & 77.25 & \underline{77.50} & 74.81 & 75.79 & 75.48 & 74.15 & \cellcolor{gray!20}\textbf{77.61}($\uparrow$+3.46) \\
Mixer-B/16 & Swin-N  & 76.62 & 75.53 & 76.26 & 75.03 & \underline{76.54} & \textbf{76.63} & 76.17 & 75.81 & 75.52 & 73.38 & \cellcolor{gray!20}75.93($\uparrow$+2.55) \\
\hline
ConvNeXt-T &  ResMLP-S12 & 82.05 & 76.65 & 76.84 & 77.23 & 77.24 & \underline{77.53} & 74.69 & 75.79 & 75.28 & 73.57 & \cellcolor{gray!20}\textbf{78.29}($\uparrow$+4,72) \\
Swin-T &  ResMLP-S12 & 81.38 & 76.65 & 76.67 & 76.99 & 77.25 & \underline{77.31} & 76.48 & 76.15 & 75.10 & 73.40 & \cellcolor{gray!20} \textbf{77.53}($\uparrow$+4.13) \\
\hline
\end{tabular}
}
\vskip -0.2in
\label{table4}
\end{table*}
\begin{table*}[t]
\renewcommand{\arraystretch}{1.2}
\centering
\caption{\textbf{Results on Object Detection.} Average Precision (AP) under different settings is reported for evaluation. “R101” denotes the use of ResNet-101 as backbone, and the same naming convention applies to other models.}
\scalebox{1.0}{
\begin{tabular}{ccc|ccccccc}
\hline
\multicolumn{3}{c}{Method} & \multicolumn{2}{|c}{mAP} & AP50 & AP75 & APl & APm & APS\\
\hline
Teacher &  \multicolumn{2}{c}{Faster R-CNN w/ R101-FPN} & \multicolumn{2}{|c}{42.04} & 62.48 & 45.88 & 54.60 & 45.55 & 25.22 \\
Student &  \multicolumn{2}{c}{ Faster R-CNN w/ R50-FPN} & \multicolumn{2}{|c}{37.93} & 58.84 & 41.05 & 49.10 & 41.14 & 22.44 \\
\multicolumn{3}{c}{ KD } & \multicolumn{2}{|c}{38.35 (+0.42)} & 59.41 & 41.71 & 49.48 & 41.80 & 22.73 \\
\multicolumn{3}{c}{ FitNet } & \multicolumn{2}{|c}{38.76 (+0.83)} & 59.62 & 41.80 & 50.70 & 42.20 & 22.32 \\
\multicolumn{3}{c}{ FIGI } & \multicolumn{2}{|c}{39.44 (+1.51)} & 60.27 & 43.04 & 51.97 & 42.51 & 22.89 \\
\multicolumn{3}{c}{ ReviewKD } & \multicolumn{2}{|c}{\textbf{40.35} (+2.42)} & \textbf{61.00} & 43.77 & 52.87 & \textbf{43.88} & \textbf{23.53} \\
\rowcolor{gray!20}
\multicolumn{3}{c}{ MSDCRD } & \multicolumn{2}{|c}{39.45 (+1.52)} & 60.19 & 42.93 & 51.54 & 42.79 & 22.92 \\
\rowcolor{gray!20}
\multicolumn{3}{c}{ MSDCRD + FIGI } & \multicolumn{2}{|c}{39.56 (+1.63)} & 60.20 & 43.89 & 51.99 & 43.08 & 22.99 \\
\rowcolor{gray!20}
\multicolumn{3}{c}{ MSDCRD + MSE } & \multicolumn{2}{|c}{\textbf{40.35} (+2.42)} & 60.89 & \textbf{43.95} & \textbf{53.47} & 43.63 & 23.14 \\
\hline
Teacher &  \multicolumn{2}{c}{Faster R-CNN w/ R50-FPN} & \multicolumn{2}{|c}{40.22} & 61.02 & 43.81 & 51.98 & 43.53 & 24.16 \\
Student &  \multicolumn{2}{c}{ Faster R-CNN w/ MV2-FPN} & \multicolumn{2}{|c}{29.47} & 48.87 & 30.90 & 38.86 & 30.77 & 16.33 \\
\multicolumn{3}{c}{ KD } & \multicolumn{2}{|c}{30.13 (+0.66)} & 50.28 & 31.35 & 39.56 & 31.91 & 16.69 \\
\multicolumn{3}{c}{FitNet } & \multicolumn{2}{|c}{30.20 (+0.73)} & 49.80 & 31.69 & 39.69 & 31.64 & 16.39 \\
\multicolumn{3}{c}{ FIGI } & \multicolumn{2}{|c}{31.16 (+1.69)} & 50.68 & 32.92 & 42.12 & 32.63 & 16.73 \\
\multicolumn{3}{c}{ ReviewKD } & \multicolumn{2}{|c}{\textbf{33.71} (+4.24)} & \textbf{53.15} & 36.13 & 46.47 & \textbf{35.81} & 16.77 \\
\rowcolor{gray!20}
\multicolumn{3}{c}{ MSDCRD } & \multicolumn{2}{|c}{32.55 (+3.08)} & 52.46 & 34.70 & 43.42 & 34.47 & 16.65\\
\rowcolor{gray!20}
\multicolumn{3}{c}{ MSDCRD + FIGI } & \multicolumn{2}{|c}{32.59 (+3.12)} & 52.51 & 34.82 & 43.64 & 34.84 & 17.04\\
\rowcolor{gray!20}
\multicolumn{3}{c}{ MSDCRD + MSE } & \multicolumn{2}{|c}{33.55 (+4.08)} & 53.10 & 35.55 & 45.76 & 35.68 & \textbf{17.50} \\
\hline
Teacher &  \multicolumn{2}{c}{RetinaNet101} & \multicolumn{2}{|c}{40.40} & 60.25 & 43.19 & 52.18 & 44.34 & 24.032 \\
Student &  \multicolumn{2}{c}{RetinaNet50} & \multicolumn{2}{|c}{36.15} & 56.03 & 38.73 & 46.95 & 40.25 & 21.37 \\
\multicolumn{3}{c}{ KD } & \multicolumn{2}{|c}{36.76 (+0.61)} & 56.60 & 39.40 & 48.17 & 40.56 & 21.87 \\
\multicolumn{3}{c}{ FitNet } & \multicolumn{2}{|c}{36.30 (+0.15)} & 55.95 & 38.95 & 47.14 & 40.32 & 20.10 \\
\multicolumn{3}{c}{ FIGI } & \multicolumn{2}{|c}{37.29 (+1.14)} & 57.13 & 40.04 & 49.71 & 41.47 & 21.01 \\
\multicolumn{3}{c}{ ReviewKD } & \multicolumn{2}{|c}{38.48 (+2.33)} & 58.22 & \textbf{41.46} & \textbf{51.15} & 42.72 & \textbf{22.67} \\
\rowcolor{gray!20}
\multicolumn{3}{c}{ MSDCRD } & \multicolumn{2}{|c}{37.74 (+1.59)} & 57.83 & 40.68 & 49.29 & 42.04 & 21.03 \\
\rowcolor{gray!20}
\multicolumn{3}{c}{ MSDCRD + FIGI} & \multicolumn{2}{|c}{37.83 (+1.68)} & 57.92 & 40.71 & 48.81 & 42.12 & 21.65 \\
\rowcolor{gray!20}
\multicolumn{3}{c}{ MSDCRD + MSE } & \multicolumn{2}{|c}{\textbf{38.49} (+2.34)} & \textbf{58.57} & 41.38 & 50.32 & \textbf{42.74} & 21.68\\
\hline
\end{tabular}
}
\vskip -0.1in
\label{table5}
\end{table*}
\textbf{Results on Homogeneous Models.}
Existing advanced distillation methods can be categorized into three groups based on the space in which distillation is performed: logits-based~\cite{hinton2015distilling,hao2023one,deng2022distpro}, attention-based~\cite{zagoruyko2016paying,guo2023class}, and feature-based~\cite{romero2014fitnets,park2019relational,tian2019contrastive,chen2022knowledge,heo2019comprehensive,chen2021distilling,liu2023norm} approaches.

Table~\ref{table1} presents the comparison of homogeneous distillation results on CIFAR-100. 
MSDCRD consistently achieves SOTA performance across a wide range of heterogeneous architectures, regardless of whether the teacher and student networks share the same architecture or adopt different styles.
Compared with the conventional CRD, MSDCRD delivers consistent improvements ranging from $1.13\%$ to $3.54\%$ across various teacher–student pairs. In several cases, such as ResNet32$\times$4→WRN-40-2 and ResNet110→ResNet32, the student models even surpass their teachers.

Table~\ref{table2} presents the comparison of homogeneous distillation results on ImageNet. The widely adopted ResNet34→ResNet18 teacher–student pair on ImageNet is employed to further verify the effectiveness of MSDCRD under homogeneous settings.
MSDCRD achieves the best performance among state-of-the-art distillation methods.

\textbf{Results on Heterogeneous Models.}
Given the substantial discrepancies in feature representations among heterogeneous models, as discussed in Section~\ref{FRG}, extensive experiments are further conducted to validate the robustness and model-agnostic property of MSDCRD.
Since many previous methods were primarily designed for homogeneous settings, their performance may degrade when applied to heterogeneous scenarios.
Therefore, the experiments include comparisons with OFA, which is specifically tailored for heterogeneous distillation, as well as several strong baselines.
Following the same categorization protocol as in the homogeneous experiments, the methods are grouped into logits-based~\cite{hinton2015distilling,deng2022distpro,zhao2022decoupled,hao2023one} and feature-based~\cite{romero2014fitnets,park2019relational,tian2019contrastive,chen2021distilling,peng2019correlation} approaches.
Specifically, comprehensive cross-architecture distillation is conducted among three heterogeneous model families: CNNs, Transformers, and MLPs, to evaluate the effectiveness of MSDCRD under heterogeneous conditions.

Tables~\ref{table3} and~\ref{table4} present the comparative results of MSDCRD on CIFAR-100 and ImageNet, respectively.
It can be observed that many feature-based distillation methods originally designed for homogeneous settings, similar to conventional CRD, struggle to bridge the substantial feature discrepancies across heterogeneous models, leading to unsatisfactory results.

Table~\ref{table3} presents the comparison of heterogeneous distillation results on CIFAR-100.
MSDCRD achieves state-of-the-art performance across all heterogeneous model combinations.
Table~\ref{table4} presents the comparison of heterogeneous distillation results on ImageNet.
MSDCRD again delivers state-of-the-art performance on the majority of heterogeneous teacher–student pairs, with improvements of $2.45\%$ to $4.72\%$ over traditional CRD.
However, when Mixer-B/16 is used as the teacher, the performance gains are relatively modest compared with those observed for other teacher–student pairs.
This may be attributed to the weaker baseline performance of Mixer-B/16 relative to other teacher models, which can limit the effectiveness of knowledge transfer.
Importantly, MSDCRD achieves these competitive results using only a single-layer feature, whereas most high-performing methods rely on multi-layer features, further underscoring its efficiency and general applicability.
\subsection{Object Detection}
\label{Object Detection}
Since most knowledge distillation methods are primarily designed for classification tasks, only a few~\cite{hinton2015distilling,romero2014fitnets} apply to object detection.
In recent years, a number of distillation methods specifically designed for object detection tasks have been proposed.
Among these methods, FIFG~\cite{wang2019distilling} is selected as a representative and widely adopted baseline. 
FIFG addresses the inherent imbalance between foreground and background regions by explicitly separating them and performing distillation exclusively on the regions of interest.

\textbf{Dataset.}
Object detection experiments are conducted on the COCO 2017 dataset~\cite{lin2014microsoft}, a large-scale benchmark widely adopted for evaluating detection and segmentation tasks. 
The dataset consists of over 118,000 training images and 5,000 validation images, annotated with bounding boxes and instance segmentation masks across 80 object categories.
Its challenging nature, characterized by dense object layouts, large intra-class variance, and scale diversity, makes it a standard benchmark for assessing the performance and generalization of detection models.

\textbf{Implementation Details.}
For model implementation and training, Detectron2~\cite{wu2019detectron2}, a high-performance object detection library developed by Facebook AI Research (FAIR), is employed.
Built on PyTorch, Detectron2 offers a modular and extensible platform for implementing state-of-the-art detection algorithms such as Faster R-CNN, RetinaNet, and Mask R-CNN.
All object detection experiments in this study are carried out using four NVIDIA 3090 GPUs.

\textbf{Results and Analysis.}
The comparison results are reported in Table~\ref{table5}.
As shown, traditional classification-based methods such as KD and FitNet offer limited improvements when directly applied to detection tasks.
FIFG, owing to its detection-aware design, achieves more substantial gains, while ReviewKD employs a review mechanism that introduces feature fusion modules to guide higher-layer student features using lower-layer teacher features, thereby attaining the best performance.
From Table~\ref{table5}, it can also be observed that applying MSDCRD alone yields suboptimal results due to the inherent imbalance between foreground and background samples.
Nevertheless, its plug-and-play nature and strong extensibility render MSDCRD highly adaptable for integration with other advanced distillation techniques, leading to further improvements in student performance.
For instance, when combined with FIFG, MSDCRD effectively enhances the latter’s performance, and when integrated with MSE, MSDCRD achieves performance on par with the best distillation baseline.

Moreover, experiments are conducted on both two-stage detectors, such as Faster R-CNN~\cite{ren2015faster}, and one-stage detectors, such as RetinaNet~\cite{lin2017focal}.
The results show that the proposed approach yields substantial performance improvements for student models across different detection paradigms.
\begin{table}[t]
\centering
\renewcommand{\arraystretch}{1.5}
\captionof{table}{\textbf{Ablation Study.} Multi-Scale Decoupling (\textit{MSD}) is decomposed into two modules: Multi-Scale Pooling (\textit{MSP}) and Sample Selection (\textit{SS}). The sample-wise contrastive loss and feature-wise contrastive loss are denoted as \textit{Sample CL} and \textit{Feature CL}, respectively.}
\scalebox{0.9}{
\setlength{\tabcolsep}{4.2mm}
\begin{tabular}{c|c|c|c|c}
\hline
 \multicolumn{2}{c|}{\textit{MSD}} & \multirow{2}{*}{\textit{Sample CL}} & \multirow{2}{*}{\textit{Feature CL}} & \multirow{2}{*}{Accuracy} \\
\cline{1-2}
\textit{MSP}       &     \textit{SS}     &    &   &  \\
\hline
                    &             &            &               & 80.83 \\
\checkmark          &             &            &               & 82.75 \\
\checkmark          & \checkmark  & \checkmark &               & 84.40 \\
\checkmark          & \checkmark  & \checkmark &  \checkmark   & \textbf{84.47} \\
\hline
\end{tabular}
}
\vskip -0.1in
\label{table6}
\end{table}
\begin{table}[t]
\centering
\renewcommand{\arraystretch}{1.5}
\captionof{table}{\textbf{The Selection of Scale Combinations.} 
Using Swin-Tiny as the teacher network and ResNet-18 as the student network.
Baseline Accuracy refers to the results obtained with InfoNCE as the sole contrastive loss function, while MSDCRD Accuracy denotes the results achieved with the proposed MSDCRD method.}
\scalebox{0.9}{
\setlength{\tabcolsep}{4mm}
\begin{tabular}{c|c|c}
\hline
 Scale         &  Baseline Accuracy & MSDCRD Accuracy\\
\hline
 \{1\}         &  80.83             & 80.13         \\
 \{2\}         &  82.14             & 80.48         \\
 \{4\}         &  82.29             & 81.82         \\
\hline
 \{1,2\}       &  82.23             & 82.15          \\
 \{1,2,4\}     &  82.75             & 84.47         \\
\hline
\end{tabular}
}
\vskip -0.2in
\label{table7}
\end{table}
\begin{figure}[ht] 
\centering
\subfigure[Vanilla KD]{
    \includegraphics[width=0.4\linewidth]{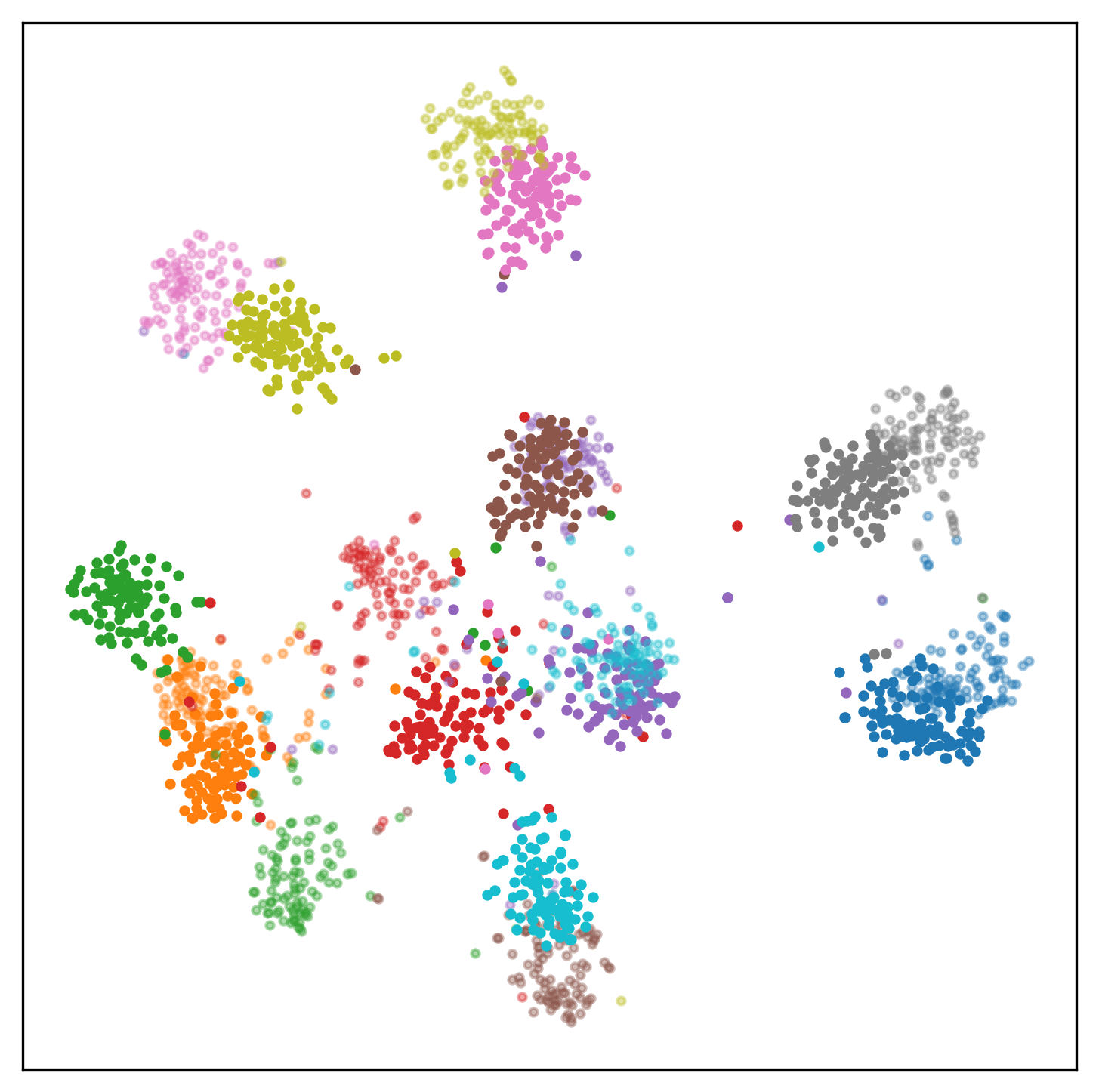}
}
\subfigure[MSDCRD]{
    \includegraphics[width=0.4\linewidth]{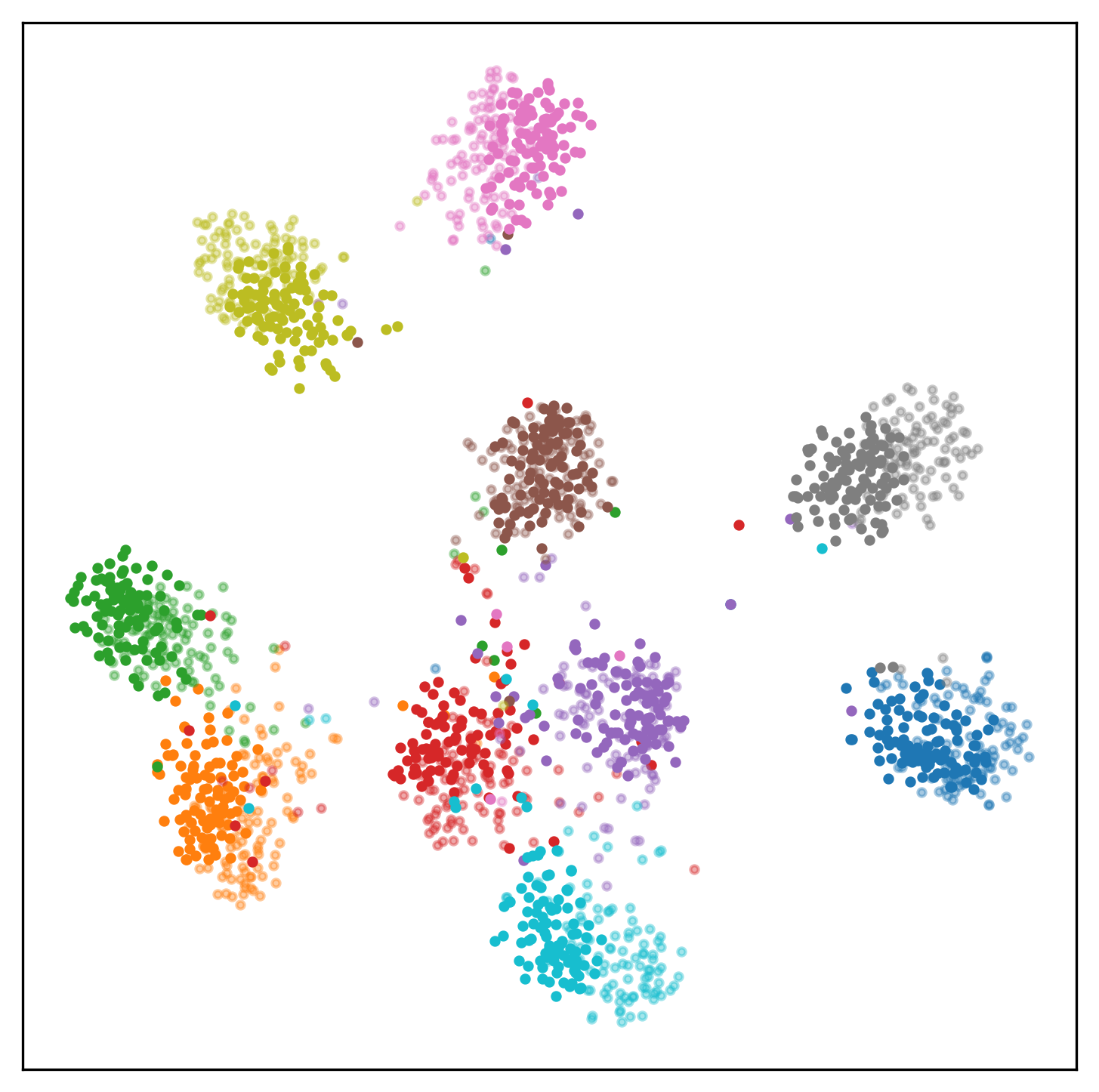}
}
\subfigure[Teacher]{
    \includegraphics[width=0.4\linewidth]{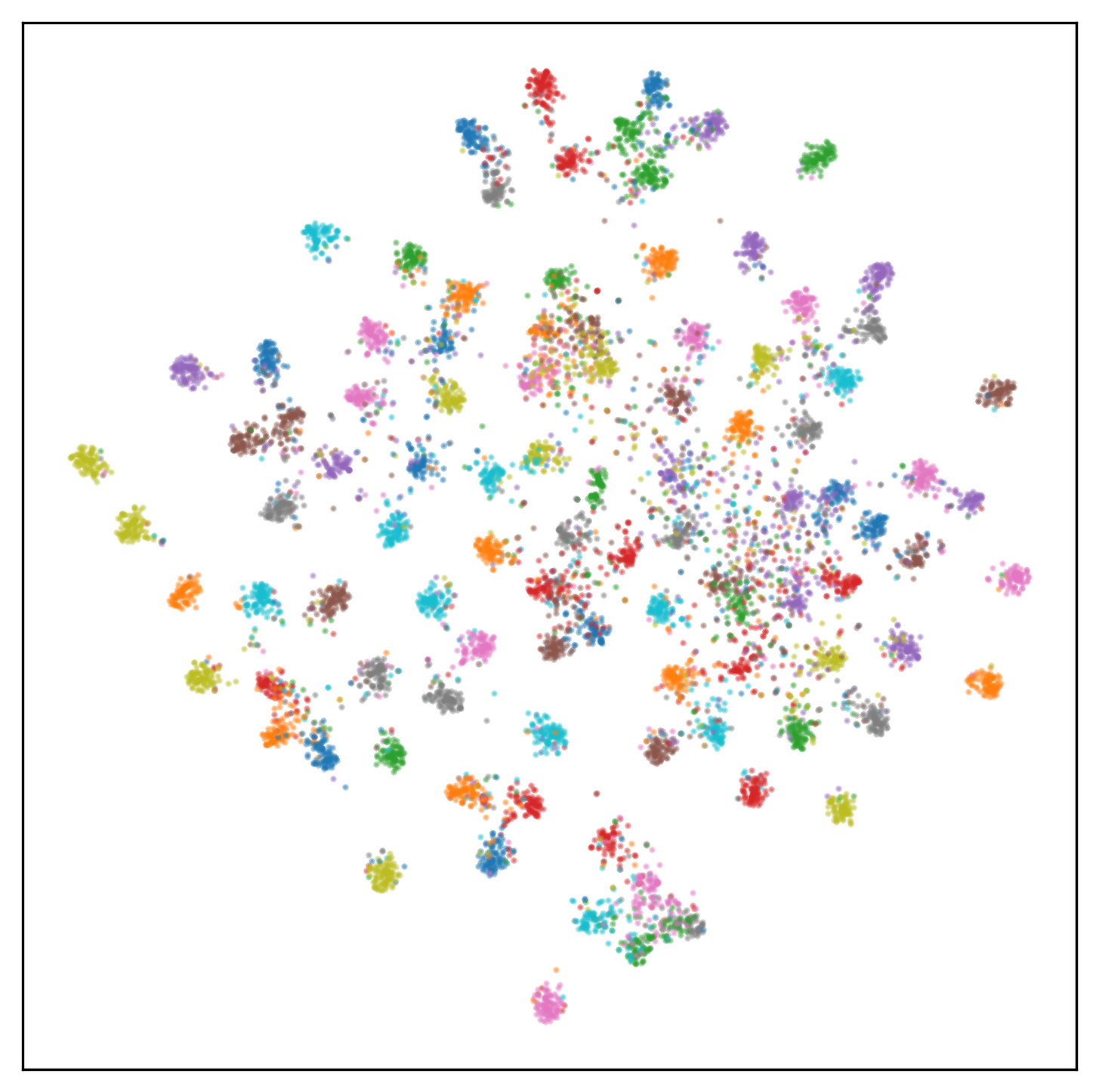}
}
\subfigure[MSDCRD]{
    \includegraphics[width=0.4\linewidth]{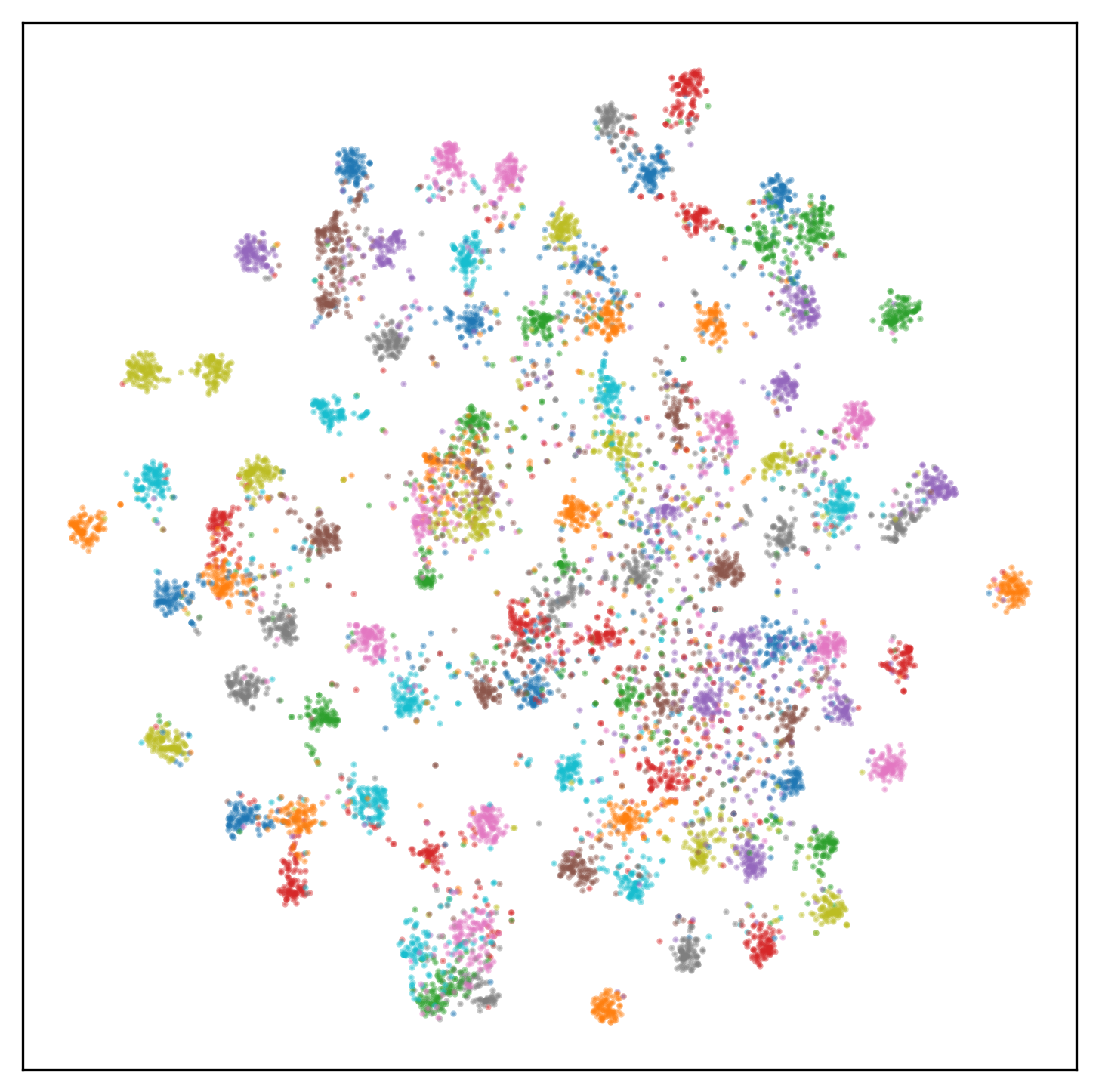}
}
\caption{\textbf{Visualization Results of Test Images from CIFAR-100 with t-SNE.} In (a) and (b), 10 classes are randomly sampled from the 100 categories. 
The features extracted by the teacher and student models are shown in dark and light colors, respectively, and under the MSDCRD method, the two distributions are nearly indistinguishable. 
In (c) and (d), the visualizations of all classes are presented for the teacher and the student trained with MSDCRD.}
\vskip -0.2in
\label{fig6}
\end{figure}
\section{Extension Experiments}
\label{Extension Experiments}
\textbf{Feature visualizations.}
As shown in Figure~\ref{fig6}(b), the features extracted by the teacher model (dark colors) and the student model distilled using MSDCRD(light colors) form tight clusters within the same class and are clearly separated across different classes, while closely aligning with the teacher network's features.

Figure~\ref{fig6}(c) and Figure~\ref{fig6}(d) visualize the feature maps of all classes from the pre-trained teacher network and the student network trained with MSDCRD, respectively. 
It is evident that MSDCRD achieves performance comparable to the teacher network.
This demonstrates the effectiveness of MSDCRD in transferring knowledge from teacher to student.

\textbf{Ablation Study.}
To assess the contribution of each component introduced in Section~\ref{Method}, an ablation study is conducted by incrementally integrating them into the baseline.
The results are summarized in Table~\ref{table6}.
Each experiment is repeated three times, and the average accuracy is reported.
The setting uses ResNet18 as the student network and Swin-T as the teacher network, with the conventional InfoNCE loss function serving as the baseline.
By introducing the \textit{MSP} module from multi-scale decoupling(\textit{MSD}), the performance of the student network already surpasses the baseline.
Incorporating the \textit{SS} module together with \textit{Sample CL} further enhances the student’s performance. 
Finally, combining with \textit{Feature CL} yields the best-performing student network.

\textbf{The Selection of Scale Combinations.}
The multi-scale feature decoupling process begins with multi-scale pooling, where sliding pooling windows of varying sizes are applied to systematically capture fine-grained local features from an individual global feature, followed by a subsequent sample selection process.
To examine the impact of different pooling window scales and to identify the optimal scale combination, comparative experiments with varying scale settings are conducted.
Since the feature map size in the penultimate layer of classification tasks is typically $7 \times 7$ or $8 \times 8$, the maximum pooling window size is set to 4.
To further ensure the generality and persuasiveness of the results, experiments with different scale combinations are carried out not only in the proposed MSDCRD method but also under the baseline setting.
As presented in Table~\ref{table7}, progressively expanding the scale combination (\{1\}, \{2\}, \{4\} → \{1,2\} → \{1,2,4\}) allows the student network to acquire increasingly rich and fine-grained knowledge within the same experimental settings, thereby yielding consistent and notable performance gains.
The best result is achieved with the scale combination \{1,2,4\}. 
These results confirm the effectiveness of the proposed design.
\section{CONCLUSION}
This work first reveals that existing feature-based distillation methods focus solely on global feature alignment, thereby overlooking the couplings among different local regions within individual global features and often causing semantic confusion.
To address this issue, a novel method, MSDCRD, is proposed, which explicitly decouples different local regions within an individual global feature and integrates them with two tailored contrastive loss functions.
This design not only eliminates the reliance of conventional CRD on large memory buffers, thereby improving efficiency, but also enables effective knowledge transfer.
MSDCRD achieves state-of-the-art results across various visual tasks in homogeneous models and further demonstrates strong knowledge transfer capability in heterogeneous settings where feature discrepancies are more pronounced, thereby highlighting its strong generalization ability.

\bibliographystyle{unsrt}
\bibliography{MSDCRD}

\newpage

 




\vfill

\end{document}